\documentclass[acmsmall, review=false, screen]{acmart}
\usepackage{booktabs} 
\usepackage[ruled]{algorithm} 
\usepackage{algorithmic}
\usepackage{amsfonts}
\usepackage{amsmath}
\usepackage{times}
\usepackage{amssymb}
\usepackage{bm}
\usepackage[skip=0pt]{caption}
\usepackage{dsfont}
\usepackage{graphicx}
\usepackage{epstopdf}
\usepackage{epsfig}
\usepackage{url}
\usepackage{subfigure}
\usepackage{color}
\usepackage{mathtools}
\usepackage{threeparttable}
\usepackage{lineno}
\usepackage{multirow}
\graphicspath{{./figures/}}
\newtheorem{theorem}{Theorem}
\DeclareMathOperator{\diag}{diag}
\newcommand{\todo}[1]{\textcolor{black}{#1}}
\acmJournal{TKDD}

\setcopyright{none}

\acmDOI{0000001.0000001}

\begin{document}
\title{Probabilistic Feature Selection and Classification Vector Machine}
\thanks{\ This work is supported by
the National Key Research and Development Program of China (Grant No. 2016YFB1000905),
the National Natural Science Foundation of China (Grant Nos. 91546116 and 91746209),
the Science and Technology Innovation Committee Foundation of Shenzhen (Grant Nos. ZDSYS201703031748284),
Ahold Delhaize,
Amsterdam Data Science,
the Bloomberg Research Grant program,
the China Scholarship Council,
the Criteo Faculty Research Award program,
Elsevier,
the European Community's Seventh Framework Programme (FP7/2007-2013) under
grant agreement nr 312827 (VOX-Pol),
the Google Faculty Research Awards program,
the Microsoft Research Ph.D.\ program,
the Netherlands Institute for Sound and Vision,
the Netherlands Organisation for Scientific Research (NWO)
under pro\-ject nrs
CI-14-25, 
652.\-002.\-001, 
612.\-001.\-551, 
652.\-001.\-003, 
and
Yandex.
All content represents the opinion of the authors, which is not necessarily shared or endorsed by their respective employers and/or sponsors.
}
\author{Bingbing Jiang}
\affiliation{
  \department{School of Computer Science and
  	Technology}
  \institution{University of Science and Technology of China}
  \city{Hefei}
  \state{Anhui}
  \postcode{230027}
  \country{China}}
  \email{jiangbb@mail.ustc.edu.cn}

\author{Chang Li}
\affiliation{
	\department{Informatics Institute}
	\institution{University of Amsterdam}
	\city{Amsterdam}
	\country{The Netherlands}}
\email{c.li@uva.nl}

\author{Maarten de Rijke}
\email{derijke@uva.nl}
\affiliation{
	\department{Informatics Institute}
	\institution{University of Amsterdam}
	\city{Amsterdam}
	\country{The Netherlands}}

\author{Xin Yao}
\affiliation{
	\department{Department of Computer Science and Engineering, Shenzhen Key Laboratory of Computational Intelligence}
	\institution{Southern University of Science and Technology}
	\city{Shenzhen}
	\state{Guangdong}
	\postcode{518055}
	\country{China}}
\email{xiny@sustc.edu.cn}

\author{Huanhuan Chen}
\authornote{Huanhuan Chen is the corresponding author.}
\affiliation{
	\department{School of Computer Science and
		Technology}
	\institution{University of Science and Technology of China}
	\city{Hefei}
	\state{Anhui}
	\postcode{230027}
	\country{China}}
\email{hchen@ustc.edu.cn}

\begin{abstract}
Sparse Bayesian learning is a state-of-the-art supervised learning algorithm that can choose
a subset of relevant samples from the input data and make reliable probabilistic predictions.
However, in the presence of high-dimensional data with irrelevant features, traditional sparse Bayesian classifiers suffer from performance degradation and low efficiency by failing to eliminate irrelevant features.
To tackle this problem, we propose a novel sparse Bayesian embedded feature selection method
that adopts truncated Gaussian distributions as both sample and feature priors.
The proposed method, called  probabilistic feature selection and classification vector machine (PFCVM$_{LP}$), is able to simultaneously select relevant features and samples for classification tasks.
In order to derive the analytical solutions, Laplace approximation is applied to compute approximate posteriors and marginal likelihoods.
Finally, parameters and hyperparameters are optimized by the type-\uppercase\expandafter{\romannumeral2} maximum likelihood method.
Experiments on three datasets validate the performance of PFCVM$_{LP}$ along two dimensions: classification performance and
effectiveness for feature selection.
Finally, we analyze the generalization performance and derive a generalization error bound for PFCVM$_{LP}$.
By tightening the bound, the importance of feature selection is demonstrated.
\end{abstract}

%
%

%
%

\keywords{Feature selection, probabilistic classification model, sparse Bayesian learning, supervised learning, EEG emotion recognition}

\maketitle


\section{Introduction}

\label{introducation}
In supervised learning, we are given input feature vectors $\mathbf{x}= \{x_{i}\in \mathbb{R}^M\}_{i=1}^N $ and corresponding labels $\mathbf{y} = \{y_{i}\}_{i=1}^N$.\footnote{In this paper, the subscript of a sample $x$, i.e., $x_{i}$, denotes the $i$-th sample and the superscript of a sample $x$, i.e., $x^{k}$, denotes the $k$-th dimension.}
The goal is to predict the label of a new datum $\hat{x}$ based on the training dataset ${\mathbf S} = \{\mathbf{x,y}\}$ together with other prior knowledge.
For regression, we are given continuous labels ${y} \in \mathbb{R}$, while for classification we are given discrete labels.
In this paper, we focus on the binary classification case, in which ${y} \in \{-1,+1\}$.

Recently, learning sparseness from large-scale datasets has generated significant research interest~\citep{svn1995,li2014,chen2014,he_fs,IRSFM}. Among the methods proposed, the support vector machine (SVM) \citep{svn1995}, which is based on the kernel trick~\citep{vapnik1998} to create a non-linear decision boundary with a small number of support vectors, is the state-of-the-art algorithm. The prediction function of SVM is a combination of basis functions:\footnote{In the rest of this paper,  we prefer to use the term of basis function instead of kernel function, because, except for SVM, the basis functions used in this paper are free of Mercer's condition.}
\begin{equation}
\label{eq:km}
f(\hat{x};\mathbf{w}) = \sum_{i=1}^N \phi(\hat{x},x_{i})w_{i} + b ,
\end{equation}
where $\phi(\cdot,\cdot)$ is the basis
function, $\mathbf{w} = \{w_i\}_{i=1}^N$ are sample weights, and $b$ is the bias.

Similar to SVM, many sparse Bayesian classifiers also use Equation \eqref{eq:km} as their decision function; examples include the relevance vector machine (RVM)~\cite{tipping2001} and the probabilistic classification vector machine (PCVM)~\cite{chen2009}.
Unlike SVM, whose weights are determined by maximizing the decision margin and limited to hard binary classification, sparse Bayesian algorithms optimize the parameters within a maximum likelihood framework and make predictions based on the average of the prediction function over the posterior of parameters. For example, PCVM computes the maximum a posteriori (MAP) estimation using the expectation-maximization (EM) algorithm; RVM and efficient probabilistic classification vector machine (EPCVM)~\citep{chen2014} compute the type-\uppercase\expandafter{\romannumeral2} maximum likelihood~\citep{berger1985} to estimate the distribution of the parameters. However, these algorithms have to deal with different scales of features due to the failure to eliminate irrelevant features.

\todo{In addition to sparse Bayesian learning, parameter-free Bayesian methods that are based on the class-conditional distributions, have been proposed to solve the classification task~\cite{smpm,kwok2007class,mpm,mempm}. \citet{mpm} proposed the minimax probability machine (MPM) to estimate the bound of classification accuracy by minimizing the worst error rate. To efficiently exploit structural information of data, \citet{smpm} proposed a structural MPM (SMPM) that can produce the non-linear decision hyperplane by using the kernel trick. To exploit structural information, SMPM adopts a clustering algorithm to detect the clusters of each class and then calculates the mean and covariance matrix for each cluster. However, selecting a proper number of clusters per class is difficult for the clustering algorithm, and calculating the mean and covariance matrix for each cluster has a high computational complexity for high-dimensional data. Therefore, SMPM cannot fit different scales of features and might suffer from the instability and low efficiency especially for high-dimensional data.}

In order to fit different scales of features, basis functions are always controlled by basis parameters (or kernel parameters).
For example, in LIBSVM~\citep{libsvm2011chang} with Gaussian radial basis functions (RBF) $\phi({x,z}) = \exp(-{\vartheta}{\|{x-z}\|^2})$, the default  $\vartheta$ is set relatively small for high-dimensional datasets and large for
low-dimensional datasets.
Although the use of basis parameters may help to address the curse of dimensionality~\citep{bellman1961adaptive},
the performance might be degraded when there are lots of irrelevant and/or redundant features~\citep{li2014,krishnapuram2004,wsvm}.
Parameterized basis functions are designed to deal with this problem.
There are two popular basis functions that can incorporate feature parameters easily:
\begin{align}
\intertext{Gaussian RBF}
\phi_{\boldsymbol{\theta}}({x,z}) & = \exp{(-\sum_{k=1}^M\theta_k(x^k -
z^k)^2) }\text{,} \\
\intertext{Pth order polynomial:}
\phi_{\boldsymbol{\theta}}({x,z}) & =(1+\sum_{k=1}^M\theta_k x^kz^k)^P \text{,}\label{eq:kernel}
\end{align}
where the subscript denotes the corresponding index of features, and $\boldsymbol \theta \in \mathbb{R}^M$ are feature parameters (also called feature weights).
Once a feature weight $\theta_k \rightarrow0$,\footnote{Practically, lots of feature weights $\theta_k \rightarrow  0$. When a certain $\theta_k$ is smaller than a threshold, we will set it $0$.} the corresponding feature will not contribute to the classification.

\todo{Feature selection, as a dimensionality reduction technique, has been extensively studied in
machine learning and data mining, and various feature selection methods have been proposed \cite{weston2000feature,li2014,wsvm,mrmr2005,trc,fsmn,l1-svm,wu2017,krishnapuram2004,yu2016,yu2015,wang2017,mohsenzadeh2013,wu2013,wang2015,yang2013,wu2010}.}
Feature selection methods can be divided into three groups: \emph{filter methods}~\citep{mrmr2005,trc,fsmn,ls2005}, \emph{wrapper methods}~\citep{weston2000feature}, and \emph{embedded methods}~\citep{li2014,l1-svm, IRSFM,wsvm,krishnapuram2004,mohsenzadeh2013}.
\todo{Filter methods independently select the subset of features from the classifier learning.
Wrapper methods consider all possible feature subsets and then select a specific subset based on its
predictive power.
Therefore, the feature selection stage and classification model are separated and independent in the filter and wrapper methods, and the wrapper methods might suffer from high computational complexity especially for high-dimensional data \cite{IRSFM}.}
Embedded methods embed feature selection in the training process, which aims to combine the advantages of the filter and wrapper methods.
\todo{
As to filter methods, \citet{mrmr2005} proposed a minimum redundancy and maximum relevance (mRMR) method, which selects relevant features and simultaneously removes redundant features according to the mutual information. To avoid evaluating the score for each feature individually like Fisher Score \cite{fscore}, a filter method, trace ratio criterion (TRC) \cite{trc} was designed to find the globally optimal feature subset by maximizing the subset level score.
Recently, sparsity regularization in feature space has been widely applied to feature selection tasks.
In \cite{l1-svm}, Bradley and Mangasarian proposed an embedded method, L$_1$SVM, that uses the L$_1$ norm to yield a sparse solution. However, the number of features selected by L$_1$SVM is upper
bounded by the number of training samples, which limits its application on high-dimensional data.
\citet{fsmn} employed joint L$_{21}$ norm minimization on both loss function and regularization to propose a filter method, FSNM.
Based on the basis functions mentioned above, \citet{wsvm} designed an embedded feature selection model, Weight SVM (WSVM), that can jointly perform feature selection and classifier construction for non-linear SVMs.
However, filter methods  are not able to adaptively select relevant features, i.e., they require a predefined number of selected features.
}

For Bayesian feature selection approaches, a joint classifier and feature optimization algorithm (JCFO) 
is proposed in \citep{krishnapuram2004}; the authors adopt a sparse Bayesian model to simultaneously perform classifier learning and feature selection.
To select relevant features, JCFO introduces hierarchical sparseness promoting priors on feature weights and then employs EM and gradient-based methods to optimize the feature weights.
In order to simultaneously select relevance samples and features, \citet{mohsenzadeh2013} extend the standard RVM and then design the relevance sample feature machine (RSFM) and an incrementally learning version (IRSFM) \citep{IRSFM}, that scales the basis parameters in RVM to a vector and applies zero-mean Gaussian priors on feature weights to generate sparsity in the feature space.
\citet{li2014} propose an EM algorithm based joint feature selection strategy for PCVM (denoted as PFCVM$_{EM}$), in which they add truncated Gaussian priors to features to enable PCVM to jointly select relevant samples and features.
However, JCFO, PFCVM$_{EM}$, and RSFM use an EM algorithm to calculate a maximum a posteriori point estimate of the sample and feature parameters.
As pointed out by \citet{chen2014}, the EM  algorithm has the following limitations: first, it is sensitive to the starting points and cannot guarantee convergence to global maxima or minima; second, the EM algorithm results in a MAP point estimate, which limits to the Bayes estimator with the 0-1 loss function and cannot represent all advantages of the Bayesian framework.

JCFO, RSFM, and IRSFM adopt a zero-mean Gaussian prior distribution over sample weights, and RSFM and IRSFM also use this prior distribution over feature weights.
As a result of adopting a zero-mean Gaussian prior over samples, some training samples that belong to the positive class ($y_i = {+1}$) will receive negative weights and vice versa; this may result in instability and  degeneration in solutions \citep{chen2009}.
Also, for RSFM and IRSFM, zero-mean Gaussian feature priors will lead to negative feature weights, which reduces the value of kernel functions for two samples when the similarity in the corresponding features is increased \citep{krishnapuram2004}.
Finally, RSFM and IRSFM have to construct an $N\times M$ kernel matrix for each sample, which yields a space complexity of at least $O(N^2M)$ to store the designed kernel matrices.

We propose a complete sparse Bayesian method, i.e., a Laplace approximation based feature selection method PCVM (PFCVM$_{LP}$), that uses the type-\uppercase\expandafter{\romannumeral2} maximum likelihood method to arrive at a fully Bayesian estimation.
In contrast to the filter methods such as mRMR~\citep{mrmr2005}, FSNM \cite{fsmn} and TRC \citep{trc}, and the embedded methods such as JCFO \cite{krishnapuram2004}, L$_1$SVM \cite{l1-svm} and WSVM \citep{wsvm}, the proposed PFCVM$_{LP}$ method can adaptively select informative and relevant samples and features with probabilistic predictions.
Moreover, PFCVM$_{LP}$ adopts truncated Gaussian priors as both sample and feature priors, which obtains
a more stable solution and avoids the negative values for sample and feature weights.
We summarize the main contributions as follows:
\begin{itemize}
    \item \todo{Unlike traditional sparse Bayesian classifiers, like PCVM and RVM, the proposed algorithm  simultaneously selects the relevant features and samples, which leads to a robust classifier for high-dimensional data sets.}

    \item \todo{Compared with PFCVM$_{EM}$~\citep{li2014}, JCFO~\citep{krishnapuram2004} and RSFM~\citep{mohsenzadeh2013}, PFCVM$_{LP}$ adopts the type-\uppercase\expandafter{\romannumeral2} maximum likelihood \citep{tipping2001} approach to approximate a fully Bayesian estimate, which achieves a more stable solution and might avoid the limitations caused by the EM algorithm.}

    \item \todo{PFCVM$_{LP}$ is extensively evaluated and compared with state-of-the-art feature selection methods on different real-world datasets. The results validate the performances of PFCVM$_{LP}$.}

    \item  We derive a generalization bound for PFCVM$_{LP}$. By analyzing the bound, we demonstrate the significance of feature selection and introduce a way of choosing the initial values.
\end{itemize}

The rest of the paper is structured as follows.
Background knowledge of sparse Bayesian learning is introduced in Section~\ref{sec:Back}. Section~\ref{sec:PFCVM} details the implementation of simultaneously optimizing sample and feature weights of PFCVM$_{LP}$.
In Section~\ref{sec:Exp} experiments are designed to evaluate both the accuracy of classification and the effectiveness of feature selection.
Analyses of sparsity and generalization for PFCVM$_{LP}$ are presented in Section~\ref{sec:bound}.
We conclude in Section~\ref{sec:conclusion}.


\section{Sparse Bayesian Learning Framework}
\label{sec:Back}
In the sparse Bayesian learning framework, we usually use the Laplace distribution and/or the student's-t distribution as the sparseness-promoting prior. In binary classification problems, we choose a Bernoulli distribution as the likelihood function. Together with the proper marginal likelihood, we can compute the parameters' distribution (posterior distribution) either by MAP point estimation or by a complete Bayesian estimation approximated by type-\uppercase\expandafter{\romannumeral2}-maximum likelihood. Below, we detail the implementation of this framework.

\subsection{Model specification}
\label{sec:Back:model}

We concentrate on a linear combination of basis functions. To simplify our notation, the decision function is defined as:
\begin{equation}
\label{eq:df}
f(\mathbf{x};\mathbf{w},\boldsymbol{\theta}) = \boldsymbol{\Phi}_{{\boldsymbol \theta}}(%
\mathbf{x})\mathbf{w} \text{,}
\end{equation}
where $\mathbf{w}$ denotes the $N+1$-dimensional sample weights; $w_0$ denotes the bias; $\boldsymbol{\Phi}_{{\boldsymbol \theta}}(\mathbf{x})$ is an $N\times(N+1)$ basis function matrix, except for the first column $\boldsymbol{\phi}_{\boldsymbol \theta ,0}(\mathbf{x})= [1,\ldots,1]^T$, other component
$\phi_{\boldsymbol \theta,ij} = \phi_{\boldsymbol{\theta}}(x_{i},x_{j})\times y_{j}$,\footnote{We assume that each sample weight has the same sign as the corresponding label.
So by multiplying the basis vector with the corresponding label, we can assume that all sample weights are non-negative.}
and $\boldsymbol \theta \in \mathbb{R}^M$ is the feature weights.

As probabilistic outputs are continuous values  in $[0,1]$, we need a link function to obtain a smooth transformation from $[-\infty,+\infty]$ to $[0,1]$.
Here, we use a sigmoid function $\sigma(z) = \frac{1}{1+e^{-z}}$ to  map Equation \eqref{eq:df} to $[0,1]$. Then, we combine this mapping with a
Bernoulli distribution to compute the following likelihood function:
\begin{equation*}  \label{eq:likelihood}
p(\mathbf{t\mid w,\boldsymbol \theta},\mathbf S) = \prod_{i=1}^{N}
\sigma_i^{t_i}(1-\sigma_i)^{(1-t_i)} \text{,}
\end{equation*}
where $t_i = (y_i+1)/2$ denotes the probabilistic target of the $i$-th sample and $\sigma_i$ denotes the sigmoid mapping for
the $i$-th sample: $\sigma_i = \sigma(f(x_i;\mathbf{w,\boldsymbol \theta}))$.
The vector $\mathbf{t} = (t_1,\ldots,t_N)^T$ consists of the probabilistic targets of all
training samples and $\mathbf{S} = \{\mathbf{x}, \mathbf{y}\}$ is the training set.

\subsection{Priors over weights and features}
\label{sec:PFCVM:samplepriors}

According to \citet{chen2009}, a truncated Gaussian prior may result in the proper sparseness to sample weights. Following this idea, we introduce a non-negative left-truncated Gaussian prior $\mathcal N_t(w_i\mid 0,\alpha_i^{-1})$ to each sample weight  $w_i$:
\begin{eqnarray}
p(w_{i}\mid \alpha _{i}) &=&\left\{
\begin{array}{ll}
2\mathcal N(w_{i}\mid 0,\alpha _{i}^{-1}) & \text{if $w_{i}\geq 0$} \\
0 & \text{otherwise}%
\end{array}
\right.  \notag \\
&=&2\mathcal N(w_{i}\mid 0,\alpha _{i}^{-1})\cdot 1_{w_{i} \geq 0}(w_{i}),
\label{eq:weight_prior}
\end{eqnarray}%
where $\alpha_i$ (precision) is a hyperparameter, which is equal to the inverse of variance,
and  $1_{x \geq 0}(x)$ is an indicator function that returns $1$ for each $x \geq 0$ and $0$ otherwise.
For the bias $w_0$, we introduce a zero-mean Gaussian prior $\mathcal N(w_0\mid 0,\alpha_0^{-1})$:
\begin{equation}
p(w_{0}\mid \alpha_{0}) = \mathcal N(w_0\mid 0,\alpha_0^{-1}).
\label{eq:bias_prior}
\end{equation}
Assuming that the sample weights are independent and identically distributed (i.i.d.), we can compute the priors over sample weights as follows:
\begin{equation}
\begin{aligned}
p(\mathbf{w|\boldsymbol \alpha})  &=\prod\limits_{i=0}^{N}p(w_{i}|\boldsymbol \alpha _{i})= \mathcal N(w_0|0,\alpha_0^{-1})\prod\limits_{i=1}^{N}\mathcal N_{t}(w_{i}|0,\alpha _{i}^{-1})%
\text{,}
\end{aligned}
\end{equation}
where $\boldsymbol{\alpha} = (\alpha_0,\ldots,\alpha_N)^T$ and $\mathcal N_t(w_i\mid 0,\alpha_i^{-1})$ denotes the left truncated Gaussian distribution.

Feature weights indicate the importance of features. For important features, the corresponding
weights are set to relatively large values and vice versa. For irrelevant and/or redundant
features, the weights are set to $0$. Following \citep{krishnapuram2004}, we should not allow negative values
for feature weights.
Based on these discussions, we introduce left truncated Gaussian priors for feature weights.
Under the i.i.d. assumption, the prior over features is computed as follows:
\begin{equation*}
p(\boldsymbol{\theta} \mid \boldsymbol{\beta}) = \prod_{k=1}^{M}p(\theta_k\mid \beta_k) =
\prod_{k=1}^{M}\mathcal N_t(\theta_k\mid 0,\beta_k^{-1}),
\end{equation*}
where $\boldsymbol{\beta} = (\beta_1,\ldots,\beta_M)^T$ are hyperparameters of feature weights.  Each prior is formalized as follows:
\begin{eqnarray}
p(\theta_k\mid \beta_k) &=&\left\{
\begin{array}{ll}
2\mathcal N(\theta_k\mid 0,\beta_k^{-1}) & \mbox {if $\theta_k \geq 0$,} \\
0 & \text{otherwise,}
\end{array}
\right.  \notag \\
&=&2\mathcal N(\theta_k\mid 0,\beta_k^{-1}) \cdot 1_{\theta_k>0}(\theta_k) \text{.}
\label{eq:feature_prior}
\end{eqnarray}
For both kinds of priors, we introduce Gamma distributions for $\alpha_i$ and $\beta_k$  as hyperpriors.
The truncated Gaussian priors will work together with the flat Gamma hyperpriors
and result in truncated hierarchical Student's-t  priors
over weights. These hierarchical priors, which are similar to Laplace priors, work as L1 regularization and lead to  sparse solutions  \citep{chen2009,krishnapuram2004}.

\subsection{Computing posteriors}

\label{sec:PFCVMem}

The posterior in a Bayesian framework contains the distribution of all parameters. Computing  parameters boils down to updating posteriors.
Having priors and likelihood,  posteriors can be computed with the following formula:
\begin{equation}
p(\mathbf{w},\boldsymbol \theta \mid \mathbf t, \boldsymbol \alpha, \boldsymbol \beta) = \frac{p(\mathbf{t\mid w},\boldsymbol{\theta}, \mathbf S)p(\mathbf{w}\mid \boldsymbol \alpha)p(%
\boldsymbol{\theta|\beta})}{p(\mathbf{t}\mid \boldsymbol \alpha,\boldsymbol \beta,\mathbf S)}.
\label{eq:posterior}
\end{equation}
%
Some methods, such as PCVM, PFCVM$_{EM}$, and JCFO, overlook information in the marginal likelihood and use the EM algorithm to obtain a MAP point estimation
of parameters. Although an efficient estimation might be obtained by the EM algorithm,
it overlooks the information in the marginal likelihood and is not regarded as a complete Bayesian estimation. Other methods,
such as RVM and EPCVM, retain the marginal likelihood. They compute the type-\uppercase\expandafter{\romannumeral2} maximum
likelihood and obtain a complete Bayesian solution.

The predicted distribution for the  new datum $\hat{x}$ is computed as follows:
\begin{equation*}
p(\hat{y}\mid \hat{x},\mathbf t, \boldsymbol \alpha,\boldsymbol \beta) = \int p(\hat{y}\mid \hat{x},\mathbf{w},\boldsymbol \theta)p(\mathbf{w}, \boldsymbol \theta \mid \mathbf t, \boldsymbol \alpha,\boldsymbol \beta)d\mathbf{w}d{\boldsymbol \theta}.
\end{equation*}
If both terms in the integral are Gaussian distributions, it is easy to compute this integral analytically.
We will detail the implementation of PFCVM$_{LP}$ in the next section.

\section{Probabilistic Feature Selection Classification Vector Machine}
\label{sec:PFCVM}

Details of computing  sample weights and sample hyperparameters were reported by \citet{chen2014}.
In this section, we mainly focus on computing parameters and hyperparameters for features.

\subsection{Approximations for posterior distributions}
\label{sec:PFCVM:lap}

\begin{figure}[th]
\begin{center}
\includegraphics[width=2.85in,height=1.9in]{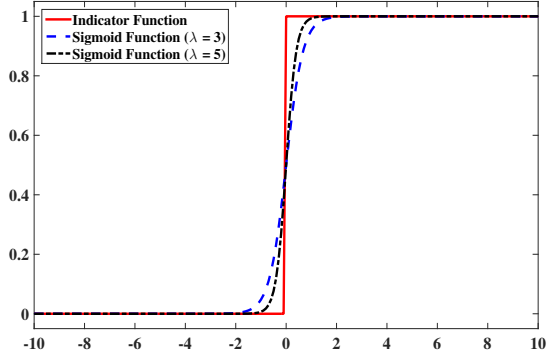}
\caption{Illustration of the indicator function and the sigmoid function.}
\label{fig:sigmoid}
\end{center}
\end{figure}

Since  the indicator function in Equation \eqref{eq:feature_prior}  is not differentiable, an approximate function is required to smoothly approximate the indicator function. Here, we use a parameterized sigmoid assumption.
Fig.\;\ref{fig:sigmoid} shows the  approximation of an indicator function made by a sigmoid function $\sigma(\lambda x)$.
As depicted in Fig.\;\ref{fig:sigmoid}, the larger $\lambda$ is, the more accurate approximation a sigmoid function will make. In PFCVM$_{LP}$, we choose $\sigma(5x)$ as the approximation function.

We calculate Equation \eqref{eq:posterior} by the Laplace approximation, in which the Gaussian distributions\footnote{Because of the
truncated prior assumption, we should take the positive quadrant part of the two Gaussian distributions, which only have an extra normalization term. Fortunately, the normalization term is independent of $\mathbf w$ and $\boldsymbol{\theta}$.
So, in the derivation, we still use the Gaussian distributions.}
$\mathcal N(\mathbf{u_{\boldsymbol \theta},\Sigma_{\boldsymbol \theta}})$ and $\mathcal N(\mathbf{u_\mathbf w,\Sigma_\mathbf w})$, are used to approximate the unknown posteriors
of feature and sample weights, respectively. We start with the logarithm of Equation (\ref{eq:posterior}) by the following formula:
\begin{equation*}
\begin{aligned}
Q(\mathbf w,\boldsymbol \theta) &=\log \{ p(\mathbf{t\mid w,\boldsymbol \theta},\mathbf S)p( \mathbf w\mid \boldsymbol \alpha)p({\boldsymbol \theta\mid \boldsymbol \beta})\} - \log p(\mathbf{t\mid \boldsymbol \alpha,\boldsymbol \beta},\mathbf S) \\
&=\sum_{n=1}^{N}\left[ t_{n}\log \sigma _{n}+(1-t_{n})\log (1-\sigma _{n})%
\right] -\frac{1}{2}{\mathbf w}^T \mathbf{A} {\mathbf w} -\frac{1}{2}{\boldsymbol \theta}^T\mathbf{B}{\boldsymbol \theta} \\
&\phantom{=}+\sum_{i=1}^{N}\log1_{w_i\geq0}(w_i)+
\sum_{k=1}^{M}\log1_{\theta_k\geq0}(\theta_k) +{const,}
\end{aligned}
\end{equation*}
where $\mathbf{A} = \diag(\alpha_0,\ldots,\alpha_N)$,
$\mathbf{B} = \diag(\beta_1,\ldots,\beta_M)$ and $const$ is independent of $\mathbf w$ and $\boldsymbol{\theta}$.

%

Using the sigmoid approximation, we substitute $1_{x\geq 0}( x)$ by $\sigma(\lambda x)$ with $\lambda=5$.
We can compute the derivative of the feature posterior function as follows:
$$\frac{\partial Q(\mathbf w,\boldsymbol \theta)}{\partial \boldsymbol{\theta}} = -\mathbf{B \boldsymbol \theta} + \mathbf{D}^T(\mathbf{t-\boldsymbol \sigma})+
\mathbf{k}_{\boldsymbol \theta} \text{,}$$
where $\mathbf{k}_{\boldsymbol \theta}=[\lambda
(1-\sigma(\lambda \theta_{1})),\ldots ,\lambda(1-\sigma (\lambda \theta_{M}))]^{T}$ is an $M$-dimensional vector,
$\boldsymbol \sigma = [\sigma_1,\ldots,\sigma_N]^T$, and
$D = \frac{\partial \boldsymbol{\Phi}_\theta \mathbf{w}}{{\partial\boldsymbol \theta}}$.
\begin{align}
\intertext{For Gaussian RBF:}
D_{i,k} & =-\sum_{j=1}^{N}w_{j}\phi _{\boldsymbol \theta,ij} (x_{i}^{k}-x_{j}^{k})^{2}.
\notag \\
\intertext{For $P$th order polynomial:}
D_{i,k} & = P x_i^{k}\sum_{j=1}^N w_j \phi_{\boldsymbol \theta,ij}^{(P-1)/P}
x_j^{k}.
\notag
\end{align}
The mean $\mathbf{u_\theta}$ of the feature posterior distribution is calculated by setting $\frac{\partial Q(\mathbf w,\boldsymbol \theta)}{\partial \boldsymbol{\theta}} = 0$:
\begin{equation}
\mathbf{u_{\boldsymbol \theta}} = \mathbf{B}^{-1}\left( \mathbf{D}^T(\mathbf{t}-\mathbf{%
\sigma})+\mathbf{k}_{\boldsymbol \theta}\right).
\label{f_means}
\end{equation}
Then we compute the second-order derivative of $Q(\mathbf w,\boldsymbol \theta)$, the Hessian matrix:
\begin{equation*}
\frac{\partial^2Q(\mathbf w,\boldsymbol \theta)}{\partial \boldsymbol{\theta}^2} = -\mathbf{O}_{\boldsymbol \theta} - \mathbf{B} -
\mathbf{D^T C D} +  \mathbf{E},
\end{equation*}
where $\mathbf{O}_{\boldsymbol \theta} = \diag(\lambda^2 \sigma(\lambda \theta_1)(1-\sigma(\lambda
\theta_1)),\ldots,\lambda^2 \sigma(\lambda \theta_M)(1-\sigma(\lambda
\theta_M)))$ is an $M\times M$ diagonal matrix, and $\mathbf{C}$ is an $N\times N$
diagonal matrix $\mathbf{C} = \diag((1-\sigma_1)\sigma_1, \ldots,(1-\sigma_N)\sigma_N)$.
$\mathbf{E}$ denotes $\frac{\partial \mathbf{D}}{\partial \boldsymbol{\theta}}^T (\mathbf{t- \sigma})$ and is computed as follows:
\begin{align}
\intertext{For Gaussian RBF:}
E_{i,k} = &\sum_{p=1}^N \Bigg [ (t_p - \sigma_p) \sum_{j=1}^{N}\phi _{\boldsymbol \theta,pj} w_{j}(x_{p}^{i}-x_{j}^{i})^{2}
\notag {} \times(x_{p}^{k}-x_{j}^{k})^{2}\Bigg ].
\notag\\
\intertext{For $P$th order polynomial:}
E_{i,k} = & \sum_{p=1}^N \Bigg [(t_p - \sigma_p) x_p^{i}x_p^{k} \sum_{j=1}^N \phi_{\boldsymbol \theta,pj}^{(P-2)/P} w_j x_j^{i} x_j^{k}\Bigg]
\notag \times P(P-1).
\notag
\end{align}
The covariance of this approximate posterior distribution equals the negative inverse of the Hessian matrix:
\begin{equation}
\boldsymbol{\Sigma_{\boldsymbol \theta}} = \left( \mathbf{D^T C D} + \mathbf{B} +\mathbf{O}_{\boldsymbol \theta}
-\mathbf{E}\right)^{-1}.  \label{f_covariance}
\end{equation}
Practically, we use Cholesky decomposition to compute the robust inversion.

In the same way, we can obtain ${\mathbf u}_\mathbf w$ and $\boldsymbol \Sigma_\mathbf w$ by computing the derivative of $Q(\mathbf w,\boldsymbol \theta)$
with respect to $\mathbf w$:
\begin{align}
\mathbf{u}_{\mathbf w} &=\mathbf{A}^{-1}\left( \boldsymbol{\Phi}^{T}_{\boldsymbol \theta}(%
\mathbf{t}-\boldsymbol{\sigma })+\mathbf{k_w}\right)
\label{s_means}
\\
\boldsymbol{\Sigma_w}&=\left(\boldsymbol{\Phi}^{T}_{\boldsymbol \theta}\mathbf{C\Phi_{\boldsymbol \theta} }+\mathbf{A}+\mathbf{O_w}\right)^{-1}\text{,}
\label{s_covariance}
\end{align}
where $\mathbf{k_w}=[0,\lambda(1-\sigma (\lambda w_{1})),\ldots ,\beta (1-\sigma (\lambda w_{N}))]^{T}$ is an $(N+1)$-dimension vector, and
$\mathbf{O_w}= \diag(0,\lambda^2 \sigma(\lambda w_1)(1-\sigma(\lambda
w_1)),\ldots,\lambda^2 \sigma(\lambda w_N)(1-\sigma(\lambda
w_N)))$ is an $(N+1)\times (N+1)$ diagonal matrix.

After the derivation, the indicator functions degenerate into vectors and matrices, $\mathbf{k}_{\boldsymbol \theta}$ in Equation \eqref{f_means}, $\mathbf{O}_{\boldsymbol \theta}$ in Equation \eqref{f_covariance} for the feature posterior, and $\mathbf{k_w}$ in Equation \eqref{s_means}, and $\mathbf{O_w}$ in Equation \eqref{s_covariance} for the sample posterior.
These two matrices will hold the non-negative property of the sample and feature weights, which is consistent with the prior assumption.

With the approximated posterior distributions,  $\mathcal N(\mathbf{u}_{\boldsymbol \theta},\boldsymbol{\Sigma}_{\boldsymbol \theta})$ and $\mathcal N(\mathbf{u}_\mathbf w,\Sigma_\mathbf w)$,
optimizing PFCVM$_{LP}$ boils down to maximizing the posterior mode of the hyperparameters, which means maximizing
$p(\boldsymbol{\alpha},\boldsymbol{\beta}\mid \mathbf{t}) \varpropto p(\mathbf{t}\mid {\boldsymbol \alpha,\boldsymbol \beta},\mathbf S) p(\boldsymbol{\alpha}) p(\boldsymbol{\beta})$
with respect to $\boldsymbol{\alpha}$ and $\boldsymbol{\beta}$. As we use flat Gamma distributions over $\boldsymbol{\alpha}$
and $\boldsymbol{\beta}$, the maximization depends on the marginal likelihood $p(\mathbf{t}\mid \boldsymbol{\alpha},\boldsymbol{\beta},\mathbf S)$ \citep{tipping2001,jiang2017}.
In the next section, the optimal marginal likelihood is obtained through the type-\uppercase\expandafter{\romannumeral2} maximum likelihood method.

\subsection{Maximum marginal likelihood}

\label{sec:PFCVM:MML}

In Bayesian models, the marginal likelihood function is computed as follows:

\begin{equation}
p(\mathbf{t}\mid \boldsymbol{\alpha},\boldsymbol{\beta},\mathbf S) = \int p(\mathbf{t}\mid \mathbf{w},\boldsymbol{\theta},\mathbf S)
p(\mathbf{w}\mid \boldsymbol{\alpha}) p(\boldsymbol{\theta}\mid \boldsymbol{\beta})d\mathbf{w}d\boldsymbol{\theta} \text{.}
\label{eq:marginal}
\end{equation}
However, when the likelihood function is a Bernoulli distribution and the priors are approximated by Gaussian distributions,
the maximization of Equation \eqref{eq:marginal} cannot be derived in closed form.
Thus we introduce an iterative estimation solution. The details of the
hyperparameter optimization and the derivation of maximizing marginal likelihood, are specified in Appendix.
Here, we use the methodology of Bayesian Occam's razor~\citep{mackay1992bayesian}.
The update formula of the feature hyperparameters is rearranged and simplified as:
\begin{align}
    \beta_k^{new} = \frac{\gamma_k}{u^2_{{\theta},k}},
\end{align}
where $u_{\boldsymbol{\theta},k}$ is the $k$-th mean of feature weights in Equation \eqref{f_means}, and we denote $\gamma_k \equiv 1- \beta_k\Sigma_{kk}$, where $\Sigma_{kk}$
is the $k$-th diagonal covariance element in Equation \eqref{f_covariance} and $\beta_k\Sigma_{kk}$ works as Occam's factor, which can automatically find a balanced solution between complexity and accuracy of PFCVM$_{LP}$. The details of updating the sample hyperparameters $\alpha_i$ are the same as for $\beta_k$, and we omit them.

In the training step, we will eliminate a feature when the corresponding $\beta_k$ is larger than a specified threshold.
In this case, the feature weight $\theta_k$ is dominated by the prior distribution and restricted to a small neighborhood around $0$.
Hence, this feature contributes little to the classification performance.
At the start of the iterative process, all samples and features are included in the model.
As iterations proceed, $N$ and $M$ are quickly reduced, which accelerates the speed of the iterations.
Further analysis of the complexity will be reported in Section \ref{sec:cc}.
In the next subsection, we demonstrate how to make predictions on new data.

\subsection{Making predictions}
\label{sec:PFCVM:MP}

When predicting the label of new sample $\hat{x}$, instead of making a hard binary decision,
we prefer to estimate the uncertainty in the decision, the posterior probability of the prediction $p(\hat{y}=1\mid \hat{x},{\mathbf S})$.
Incorporating the Bernoulli likelihood, the Bayesian model enables the sigmoid function $\sigma(f(\hat{x}))$ to
be regarded as a consistent estimate of $p(\hat{y}=1\mid \hat{x},{\mathbf S})$ \citep{tipping2001}.
We can compute the probability of prediction in the following way:
\begin{equation*}
p(\hat{y}=1\mid \hat{x},\mathbf S) =\int p(\hat{y}=1\mid \mathbf w,\hat x,\mathbf S)q(\mathbf w)d\mathbf w\text{,}
\end{equation*}
where $ p(\hat{y}=1\mid \mathbf w,\hat x,\mathbf S) = \sigma(\mathbf{u}^T_{\mathbf w}\boldsymbol{\phi}_{\boldsymbol \theta}(\hat x))$ and $q(\mathbf w)$ denotes the posterior of sample weights. Employing the posterior approximation in Section \ref{sec:PFCVM:lap}, we have $q(\mathbf w)\approx \mathcal N(\mathbf w\mid \mathbf{u_w},\boldsymbol \Sigma_{\mathbf w})$. According to \citep{bishop2006}, we have:
\begin{align*}
p(\hat{y}=1\mid \hat{x},\mathbf S)&=\int\sigma(\boldsymbol{\phi}^T_{\boldsymbol \theta}(\hat x)\mathbf{u_w}) N(\mathbf w\mid \mathbf{u_w},\boldsymbol \Sigma_{\mathbf w}) d\mathbf w \approx \sigma\big(\kappa(\sigma^2_{\hat x})\mathbf{u}^T_{\mathbf w}\phi_{\boldsymbol \theta}(\hat x)\big)\text{,}
\end{align*}
where $\kappa(\sigma^2_{\hat x})= \big(1+\dfrac{\pi}{8}\boldsymbol{\phi}^T_{\boldsymbol \theta}(\hat x) \boldsymbol \Sigma_{\mathbf w}\boldsymbol{\phi}_{\boldsymbol \theta}(\hat x)\big)^{-1/2}$ is the variance of $\hat x$ with the covariance of sample posterior distribution $\boldsymbol \Sigma_{\mathbf w}$.

To arrive at a binary classification, we choose  $\mathbf{u}^T_{\mathbf w}\boldsymbol{\phi}_{\boldsymbol \theta}(\hat x)=0$ as the decision boundary, where
we have the probability $p(\hat{y}=1\mid \hat{x},\mathbf S)=0.5$. Thus, computing the sign of  $\mathbf{u}^T_{\mathbf w}\boldsymbol{\phi}_{\boldsymbol \theta}(\hat x)$ will meet the case of  0-1 classification. Moreover, the likelihood of prediction provides the confidence of the prediction, which is more important in unbalanced classification tasks.

\subsection{Implementation}
We detail the implementation of PFCVM$_{LP}$ step by step and provide pseudo-code in Algorithm~\ref{alg:1}.
\begin{algorithm}[h]
\begin{center}
\caption{PFCVM$_{LP}$ algorithm}\label{alg:1}
\begin{algorithmic}[1]
   \STATE \textbf{Input:} Training data set: ${\mathbf S}$; initial values: \emph{INITVALUES};
   threshold: \emph{THRESHOLD}; \\
  \ \ \ \ \ \ \  \ \  \ \  \ the maximum number of iterations: \emph{maxIts}.
  \STATE  \textbf{Output:} Weights of model: $\mathbf{WEIGHT}$; \ \ \ \
    Hyperparameters: $\mathbf{HYPERPARAMETER}$.
    \STATE \textbf{Initialization:} $[\mathbf{w,\boldsymbol \theta,\boldsymbol \alpha,\boldsymbol \beta}]$= \emph{INITVALUES}; \ \ \ \ $\mathbf{Index}$ = generateIndex(${\boldsymbol \alpha,\boldsymbol \beta}$) \;
    \WHILE {$i < \mathit{maxItes}$ }
      \STATE  $\boldsymbol{\Phi} = \mathit{updateBasisFunction}(\mathbf{x,\boldsymbol \theta,Index}$) \;
        \STATE [$\mathbf{w,\boldsymbol \theta}$] = updatePosterior($\boldsymbol{\Phi}$,$\mathbf{w,\boldsymbol \theta,\boldsymbol \alpha,\boldsymbol \beta,Y}$) \;
       \STATE [${\boldsymbol \alpha,\boldsymbol \beta}$] = maximumMarginal($\boldsymbol{\Phi}$,$\mathbf{w,\boldsymbol \theta,\boldsymbol \alpha,\boldsymbol \beta,Y}$) \;
        \IF {$\boldsymbol \alpha_i$ or $\boldsymbol \beta_k >\mathit{THRESHOLD}.\mathit{maximum}$ }
          \STATE      delete the $i$th sample or the $k$th feature\;
        \ENDIF
       \STATE $\mathbf{Index}$ = updateIndex($\boldsymbol \alpha,\boldsymbol \beta$) \;
       \STATE $\mathit{marginal}$  = calculateMarginal($\boldsymbol{\Phi}$,$\mathbf{w,\boldsymbol \theta,\boldsymbol \alpha,\boldsymbol \beta,Y}$)\;

        \IF{$\Delta \mathit{marginal} < \mathit{THRESHOLD}.\mathit{minimal}$}
          \STATE  break\;
        \ENDIF
   \STATE $\mathbf{WEIGHT}= [\mathbf{w,\boldsymbol \theta,Index}]$\;
   \STATE $\mathbf{HYPERPARAMETER} = [\boldsymbol \alpha,\boldsymbol \beta]$ \;
\ENDWHILE

\end{algorithmic}
\end{center}
\end{algorithm}

  Algorithm \ref{alg:1} consists of the following main steps.
\begin{enumerate}
  \item First, the values of $\mathbf{w,\boldsymbol \theta,\boldsymbol \alpha,\boldsymbol \beta}$ are initialized by  \emph{INITVALUES} and a parameter $\mathbf{Index}$ generated to indicate the useful samples and features (line $3$).
  \item At the beginning of each iteration, compute the matrix $\boldsymbol{\Phi}$ according to Equation \eqref{eq:kernel}  (line $5$).
  \item Based on Equation \eqref{eq:posterior}, use the new hyperparameters to re-estimate the posterior (line $6$).
  \item Use the re-estimated parameters to maximize the logarithm of marginal likelihood and update the hyperparameters according to Equation \eqref{eq:marginal} (line $7$).
  \item Prune irrelevant samples and useless features if the corresponding hyperparameters are larger than a specified threshold (lines $8$, $9$, $10 $).
  \item Update the $\mathbf{Index}$ vector (line 11).
  \item Calculate the logarithm of the marginal likelihood (line $12$).
  \item Convergence detection, if the change of marginal likelihood is relatively small, halt the iteration (lines $13$, $14$, $15$).
  \item Generate the output values. The vector $\mathbf{WEIGHT}$ consists of sample and feature weights and the vector $\mathbf{Index}$ indicates the relevant
   samples and features (lines $16$, $17$).
\end{enumerate}

\noindent%
We have now presented all details of PFCVM$_{LP}$, including derivations of equations and pseudo-code.
Next, we evaluate the performance of PFCVM$_{LP}$ by comparing with other state-of-the-art algorithms on a Waveform (UCI) dataset, EEG emotion recognition datasets, and high-dimensional gene expression datasets.

\section{Experimental Results}
\label{sec:Exp}

In a series of experiments, we assess the performance of PFCVM$_{LP}$.
The first experiment aims to evaluate the robustness and stability of PFCVM$_{LP}$ against noise features.
Second, a set of experiments are carried out on the emotional EEG datasets to assess the performance of classification and feature selection.
Then, experiments are designed on gene expression datasets, which contain lots of irrelevant features.
Finally, the computational and space complexity of PFCVM$_{LP}$ is analyzed.

\subsection{Waveform dataset: Stability and robustness against noise}
\label{sec:Exp:wave}
\todo{The Waveform dataset \cite{UCI} contains a number of noise features and has been used to estimate the robustness of feature selection methods. This dataset contains 5,000 samples with 3 classes of waves (about $33\%$ for each wave). Each sample has 40 continuous features, in which the first 21 features are relevant
for classification, whereas the latter 19 features are irrelevant noise with mean 0 and variance 1. The presence of 19 noise
features in the Waveform dataset increases the hardness of the classification problem. Ideal feature selection
methods should select the relevant features (features 1--21) and simultaneously remove the irrelevant noise features
(features 22--40).}
To evaluate the stability and robustness of feature selection of PFCVM$_{LP}$ with noise features,
we choose wave $1$ vs.\ wave $2$ from the Waveform as the experimental data, which includes 3,345 samples.
In the experiment, we randomly sample data examples to generate $100$ distinct training and testing sets,
in which each training set includes $200$ training samples for each class.
Then we run PFCVM$_{LP}$ and three embedded feature selection algorithms on each data partition.

First, to compare the stability of PFCVM$_{LP}$ against that of other algorithms, two indicators are employed to measure the stability, i.e., the popular Jaccard index stability~\citep{jaccard} and the recently proposed Pearson's correlation coefficient stability~\citep{pearson}. The stability in the output feature subsets is a key evaluation metric for feature selection algorithms, which quantizes the sensitivity of a feature selection procedure with different training sets. Assume $\mathcal{F}$ denotes the set of selected feature subsets, $\bm{s_i}$, $\bm{s_j}\in \mathcal{F}$ are two selected feature subsets. The \emph{Jaccard index} between $\bm{s_i}$, $\bm{s_j}$ is defined as:
\begin{equation}
\label{jaccard_index}
  \psi_{jaccard}(\bm{s_i}, \bm{s_j})=\frac{\vert \bm{s_i}\cap\bm{s_j}\vert}{\vert\bm{s_i}\cup\bm{s_j}\vert}=\frac{r_{ij}}{r_i+r_j-r_{ij}},
\end{equation}
where $r_{ij}$ denotes the number of common features in $\bm{s_i}$ and $\bm{s_j}$, and $r_{i}$ is the size of selected features in $\bm{s_i}$. Based on the Jaccard index in Equation \eqref{jaccard_index}, the \emph{Jaccard stability} of $\mathcal{F}$ is computed as follows:
\begin{equation}
   \label{jaccard_stability}
  \Psi_{jaccard}(\mathcal{F})=\frac{2}{R(R-1)}\sum_{i=1}^{R-1}\sum_{j>i}^R\psi_{jaccard}(\bm{s_i}, \bm{s_j}),
\end{equation}
in which $R$ denotes the number of the selected feature in $\mathcal{F}$. $ \Psi_{jaccard}(\mathcal{F})\in [0,1]$, where $0$ means there is no
overlap between any two feature subsets,  $1$ means that all feature subsets in $\mathcal{F}$ are identical.

Following~\citep{pearson}, the Pearson's coefficient between $\bm{s_i}$ and $\bm{s_j}$ can be redefined as follows:
\begin{equation}
\label{pearson_index}
  \psi_{pearson}(\bm{s_i}, \bm{s_j})==\frac{M\cdot r_{ij}-r_i\cdot r_j}{\sqrt{r_i\cdot r_j(M-r_i) \cdot(M-r_j)}},
\end{equation}
where $M$ is the number of sample features. Using Equation \eqref{pearson_index}, the Pearson's correlation coefficient stability value of $\mathcal{F}$ is computed as follows:
\begin{equation}
 \label{pearson_stability}
\Psi_{pearson}(\mathcal{F})=\frac{2}{R(R-1)}\sum_{i=1}^{R-1}\sum_{j>i}^R\psi_{pearson}(\bm{s_i}, \bm{s_j}).
\end{equation}
$ \Psi_{pearson}(\mathcal{F})\in [-1,1]$, in which $-1$ means that any two feature subsets are complementary, $0$ means there is no correlation between any two feature subsets, and $1$ means that
all feature subsets in $\mathcal{F}$ are fully correlated.

\begin{figure}[ht]
	\centering
	\includegraphics[height=4cm]{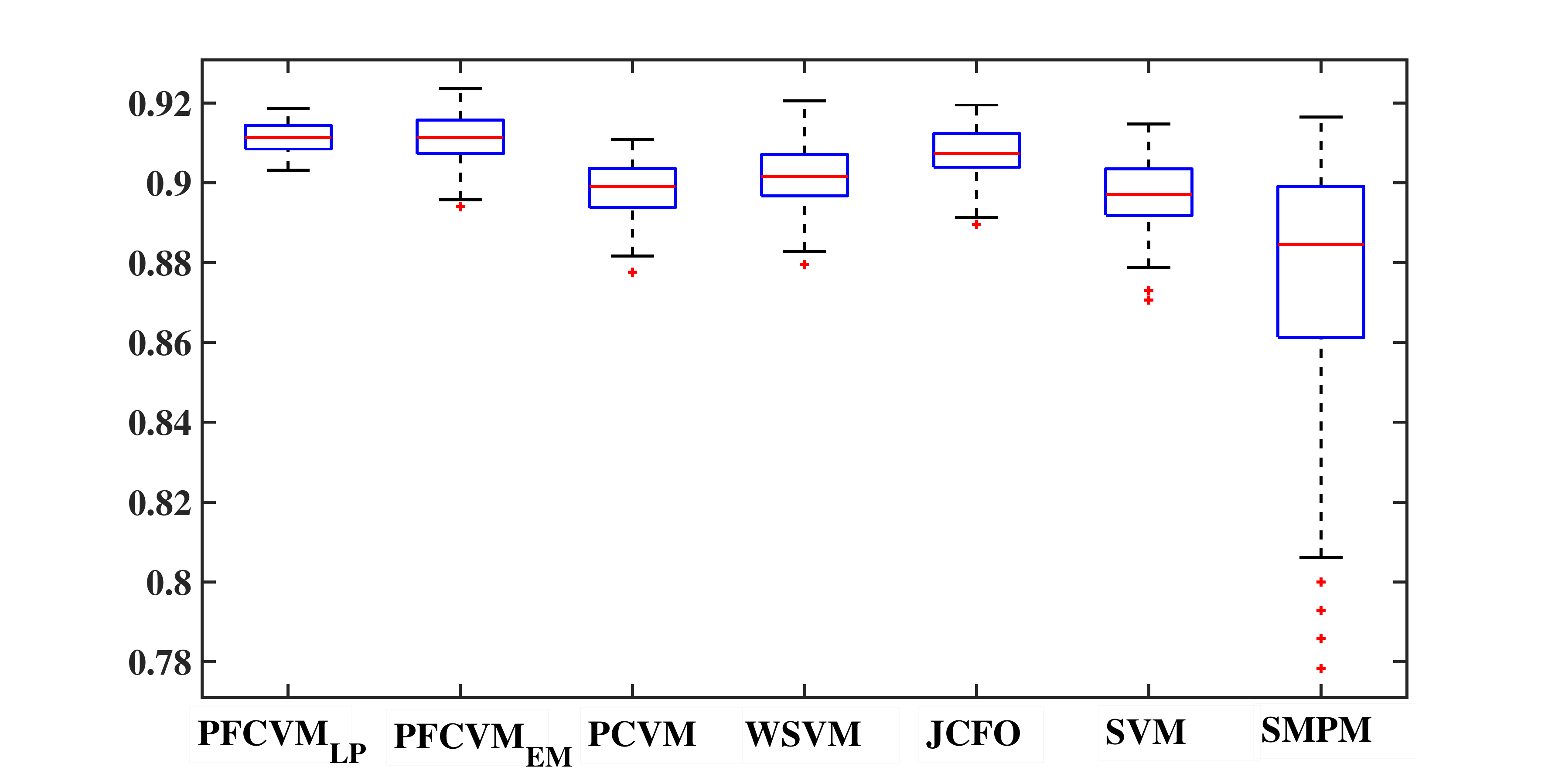}
	\caption{Classification accuracy of PFCVM$_{LP}$ and compared algorithms.}
	\label{box_figure}
\end{figure}

In order to provide comprehensive results, three embedded feature method, WSVM, JCFO, and PFCVM$_\mathit{EM}$, and three
supervised learning methods, SVM, PCVM and SMPM \cite{smpm} using all features are chosen for comparison.
The experimental settings are the same as those in \citep{chen2009}.
The experiments are repeated $100$ times with different training and test sets, and $100$ feature subsets will be obtained.
Therefore, the stability performance of each method, measured by Jaccard index and Person's correlation coefficient, is listed in Table \ref{stability},
and the classification accuracy is depicted in Fig. \ref{box_figure}.

\begin{table}[ht]
	\caption{The Jaccard and Pearson stability performances of PFCVM$_{LP}$ and other embedded feature selection algorithms on Waveform dataset.}
	\label{stability}\centering
	\resizebox {4in}{!}{
		\setlength{\tabcolsep}{0.07in}
		\begin{tabular}{ccccc}
			\toprule
			\rule{0pt}{0.3cm} Algorithms  & PFCVM$_{LP}$  & PFCVM$_{EM}$ & WSVM & JCFO  \\
			\midrule
			Jaccard & \textbf{0.556}$\pm$0.071 & 0.525$\pm$0.082 & 0.543$\pm$0.076 &  0.518$\pm$0.072     \\
			Pearson  & \textbf{0.662}$\pm$0.016 & 0.610$\pm$0.026 & 0.646$\pm$0.019 &  0.603$\pm$0.017   \\
			\bottomrule
	\end{tabular}}
\end{table}

\begin{figure*}[ht]
\centering
\includegraphics[width=\textwidth]{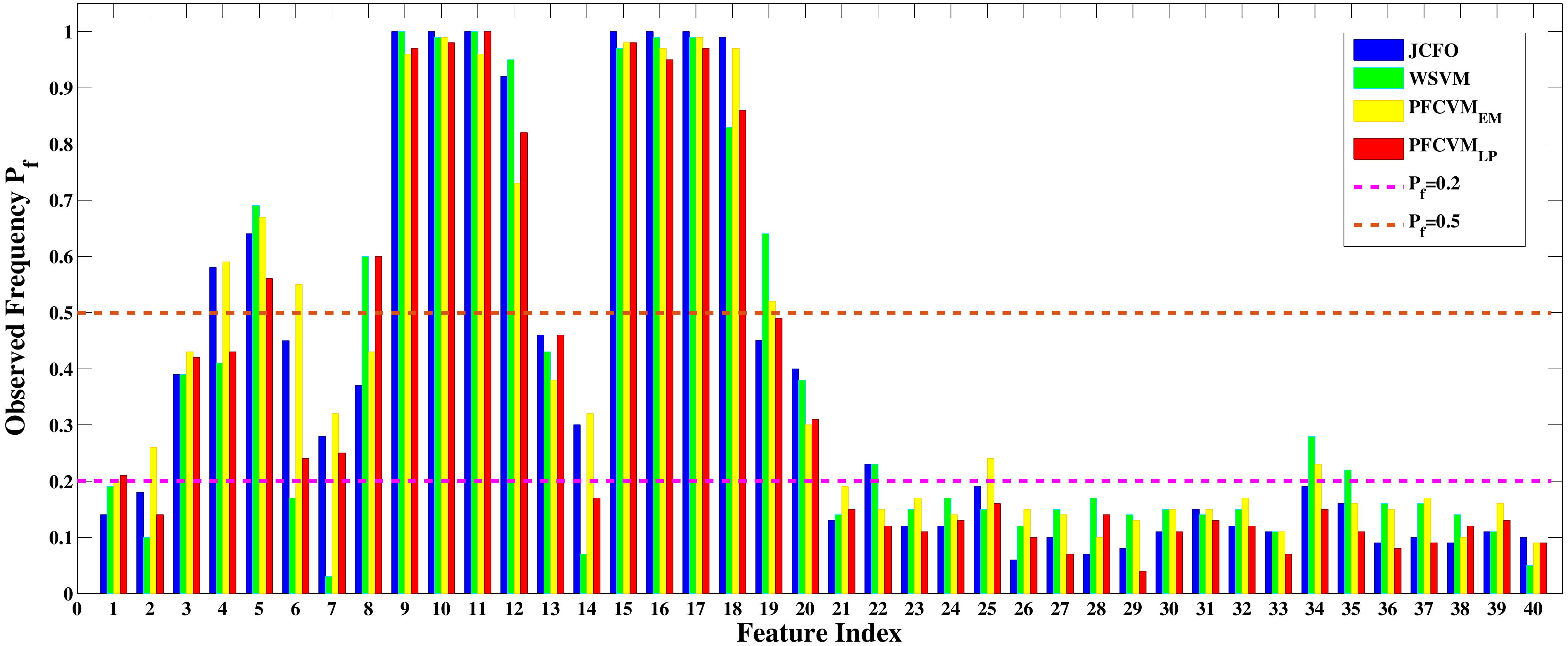}
\caption{The selected frequency of each feature. The first 21 features are actual features, and the latter 19 features are noise.}
\label{frequency_figure}
\end{figure*}
According to the stability definition in Equations (\ref{jaccard_stability}) and (\ref{pearson_stability}), a high value of stability means that the selected feature subsets do not significantly change with different
training sets. 
From Table~\ref{stability} and Fig.~\ref{box_figure}, we observe that PFCVM$_{LP}$ achieves the best stability performance in terms of both Jaccard and
Pearson index, and highly competitive accuracy in comparison with other algorithms.
The stability of WSVM is better than that of PFCVM$_\mathit{EM}$ and JCFO, which is attributed to the use of the LIBSVM \cite{libsvm2011chang} and
CVX \cite{grant2008cvx} optimization toolbox.
However, PFCVM$_\mathit{EM}$ and JCFO show inferior stability scores, the reason being that they use the EM algorithm to a point estimate of feature parameters, which suffers from the initialization
and may converge to a local optimum~\citep{chen2014}.
\todo{Finally, in Fig.~\ref{box_figure} we also note that due to the lack of feature selection, SVM and SMPM perform poorly.}

In order to demonstrate the robustness of PFCVM$_\mathit{LP}$ against the irrelevant noise features,
the selected frequency of each feature, $\hat P_f$, is shown in Fig.~\ref{frequency_figure}. From Fig.~\ref{frequency_figure},
we observe that PFCVM$_\mathit{LP}$ shows comparative effectiveness to WSVM, JCFO and PFCVM$_\mathit{EM}$ on the first 21 actual features,
and the $\hat P_f$ of features $5$, $9$, $10$, $11$, $12$, $15$, $16$, $17$, $18$ are greater than $0.5$.
\todo{As shown in Fig.~\ref{box_figure}, using these features, SVM achieves a $92.12\%$ accuracy, an improvement over the result obtained by using all features.
To quantitatively evaluate the capability of eliminating noise features for these embedded feature selection methods, the frequency of selecting the latter 19 noise features is used.
From Fig.~\ref{frequency_figure}, we note that the frequencies of selecting the
noise features are all less than $0.2$ in PFCVM$_\mathit{LP}$. However, there are $3$, $1$, $2$ noise features with more than 0.2 selected
frequencies for WSVM, JCFO, and PFCVM$_\mathit{EM}$, respectively.
This result demonstrates that in the presence of noise features, PFCVM$_\mathit{LP}$ performs much better than other algorithms in terms of eliminating those noise features.}


\subsection{Emotional EEG datasets: Emotion recognition and effectiveness for feature selection}
\label{eeg_section}

\begin{figure}[ht]
  \centering
  \includegraphics[width=.3\columnwidth]{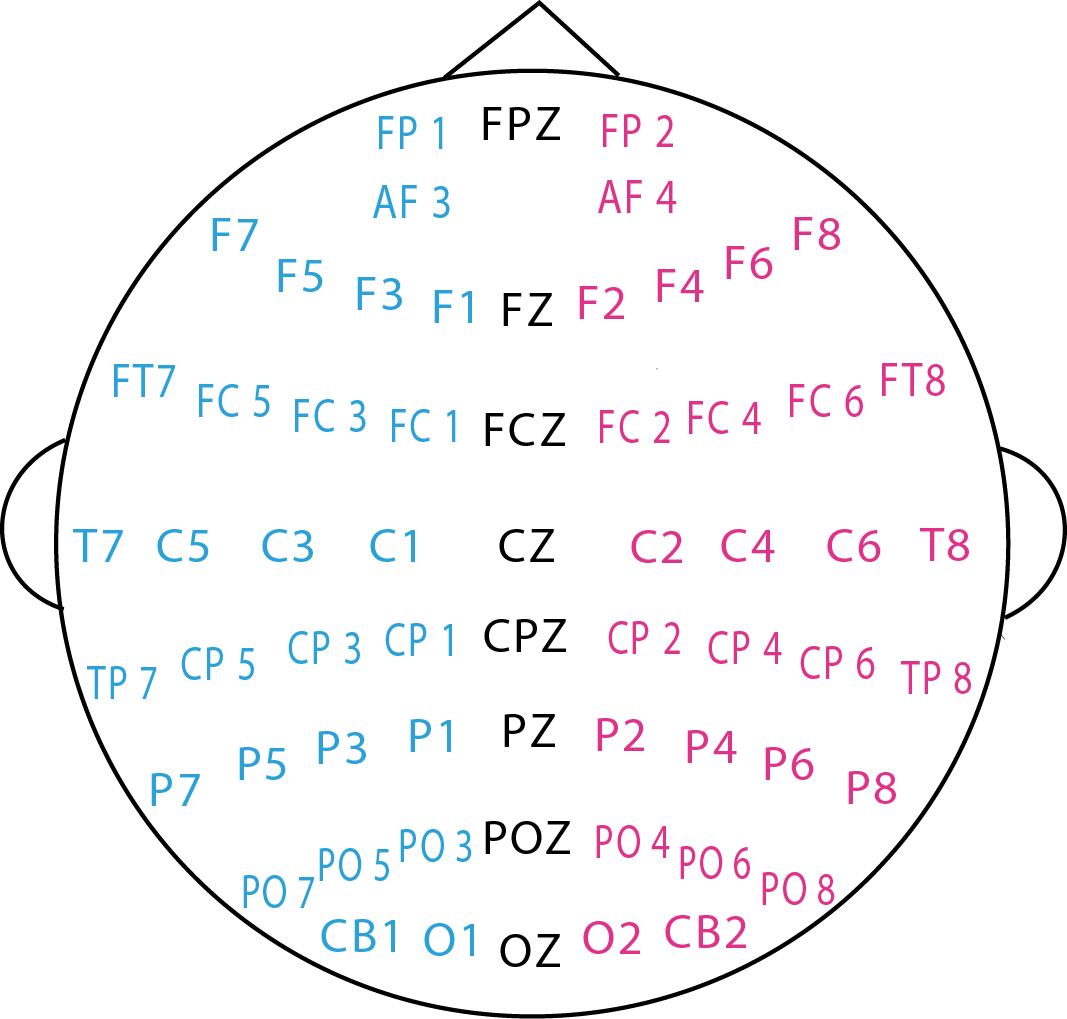}\\
  \caption{The layout of 62-channel symmetrical electrodes on the EEG.  }\label{channels}
\end{figure}

In this section, a newly developed emotion EEG dataset, SEED \cite{zheng2015}, will be used to evaluate the performance of PFCVM$_\mathit{LP}$. 
The SEED dataset contains the EEG signals of 15 subjects, which were recorded while the subjects were watching 15 emotional film clips in the emotion
experiment.
The subjects' emotional reactions to the film clips are used as the emotional labels ($-1$ for negative, $0$ for neutral and $+1$ for positive) of the
corresponding film clips.
The EEG signals were recorded by 62-channel symmetrical electrodes which are shown in Fig.~\ref{channels}. In our experiments, the differential
entropy (DE) features are chosen for emotion recognition due to its better discrimination \cite{duan2013}.
The DE features are extracted from 5 common frequency bands, namely Delta (1-3Hz), Theta (4-7Hz), Alpha (8-13Hz), Beta (14-30Hz), and Gamma (31-50Hz).
Therefore, each frequency band has 62-channel symmetrical electrodes and there are totally 310 features for one sample.
In order to investigate neural signatures and stable patterns across sessions and individuals, each subject performed the emotion experiment in three
separate sessions with an interval of about one week or longer.

In this paper, we choose \emph{positive} vs.\ \emph{negative} samples from SEED as our experimental data,
which includes the signals of five positive and five negative film clips. In this experiment,
the EEG signals recorded from one subject is regarded as one dataset,
and thus there are 15 datasets for 15 subjects \citep{zheng2017}.
Each dataset has three sessions data,
and each session contains $2,290$ samples ($1,120$ negative samples and $1,170$ positive samples) with 310 features.
For each data, we choose $1,376$ samples as the training set (recorded from three positive and three negative film clips),
the remaining in the same session as a test set. We compare the emotion recognition and feature selection effectiveness
of PFCVM$_\mathit{LP}$ with other algorithms on the testing data.

\begin{table*}[t]
	\caption{The error rate and AUC (in \%) of PFCVM$_{LP}$ compared to other algorithms on the emotional EEG
datasets. The best result for each dataset is illustrated in boldface.}
	\label{eeg_classification}\centering
	\resizebox {5.4in }{!}{
		\setlength{\tabcolsep}{0.08in}
		\begin{tabular}{ccccccccccccc}
			\toprule
\multirow{2}[4]{*}{Subject} & \multicolumn{12}{c}{    The error rate (in \%) of PFCVM$_{LP}$ and other algorithms} \\
\cmidrule{2-13} & SVM & SMPM & RVM & PCVM & mRMR & TRC & FSNM & L$_1$SVM & WSVM & PFCVM$_{EM}$ & JCFO &  PFCVM$_{LP}$    \\
			\midrule
\#1 &    	9.84	&	8.35	&	6.02	&	8.72	&	4.70 &	\textbf{4.52} 	&	4.74 	&	8.53
& 	5.80	&	6.85	&	12.18	&	4.63 	\\

\#2 &    	13.60	&	23.30	&	12.78	&	13.42	&	20.05	&	22.43 	&	19.66 	&	18.09 	&
	13.38	&	17.72	&	19.11	&	\textbf{11.85} 	\\

\#3 &    	5.03	&	7.33	&	8.28	&	9.12	&	9.70	&	6.35 	&	3.14 	&	8.77 	&
	8.17	&	4.96	&	4.60	&	\textbf{0.00} 	\\

\#4 &   	19.74	&	20.45	&	18.07	&	19.77	&	17.02	&	17.68 	&	16.65 	&	17.28 	&
	16.45	&	17.43	&	19.31	&	\textbf{16.12} 	\\

\#5 &    	18.23	&	20.93	&	19.82	&	17.98	&	16.93	&	15.61 	&	20.81 	&	18.05 	&
	\textbf{13.57} 	&	19.32	&	20.81	&	14.81	\\

\#6 &    	12.31	&	14.30	&	9.09	&	9.22	&	7.95	&	10.44 	&	10.83 	&	16.45 	&
	15.75	&	8.01	&	11.77	& \textbf{6.16} 	\\
	
\#7 &    	10.69	&	10.20	&	11.34	&	7.91	&	10.81	&	9.70 	&	14.26 	&	20.64 	&
	13.13	&	\textbf{7.48} 	&	15.97	&8.04 	\\	

\#8 &    	8.90	&	5.93	&	8.93	&	7.35	&	8.75	&	4.47 	&	4.52 	&	3.00 	&
	5.29	&	5.85	&	\textbf{0.00} 	&	2.15	\\

\#9 &   	14.85	&	11.95	&	13.72	&	12.61	&	13.15	&	14.70 	&	10.07 	&	8.28 	&
	\textbf{7.70 }	&	16.30	&	11.27	&	11.52	\\

\#10 &  	14.33	&	7.04	&	12.50	&	11.59	&	8.49	&	10.89 	&	9.23 	&	9.34 	&
	16.74	&	6.55	&	5.27	&	\textbf{3.06} 	\\

\#11 &  	12.21	&	13.72	&	12.65	&	8.45	&	7.09	&	13.13 	&	13.57 	&	\textbf{1.46} 	&
8.72	&	1.93	&	2.04	&	1.98	\\

\#12 &  	14.81	&	12.74	&	10.95	&	12.69	&	12.06	&	10.69 	&	11.71 	&	6.08 	&
	9.32	&	12.15	&	\textbf{5.58} 	&	7.99	\\

\#13 &   	10.83	&	4.89	&	10.37	&	7.17	&	3.28	&	2.74 	&	\textbf{2.12} 	&	2.42 	&
	{2.44 }	&	8.53	&	4.08	&	2.48	\\

\#14 &   	15.46	&	14.64 &	15.40	&	18.78	&	17.79	&	15.43 	&	19.69 	&	13.82 	&
	19.77	&	\textbf{12.73}	&	13.62	&	13.27	\\
Average &  12.92 &	12.56 	& 12.14 &	11.77 	& 11.27 & 	11.34 &	11.50 &	10.87 &	11.16 &	10.42 &	10.40 &	\textbf{7.43}
                                                                   \\
\midrule
\multirow{2}[4]{*}{Subject} & \multicolumn{12}{c}{The AUC (in \%) of PFCVM$_{LP}$ and other algorithms} \\									
\cmidrule{2-13}   & SVM & SMPM & RVM & PCVM & mRMR & TRC & FSNM & L$_1$SVM & WSVM &   PFCVM$_{EM}$ & JCFO & PFCVM$_{LP}$    \\
\midrule
\#1 &    	98.37	&	97.47 	&	98.56	&	97.54	&	99.64 	&	\textbf{99.70} 	    &	99.18 	&	96.42 	&	98.04	&	96.29	&	94.88	&	99.54	\\
\#2 &    	95.54	&	86.79 	&	98.14	&	99.07	&	93.22	&	92.46 	&	92.47 	&	94.98 	&	97.09	&	94.01	&	92.97	&	\textbf{99.32} 	\\
\#3 &    	98.35	&	97.75 	&	96.92	&	99.84	&	94.15	&	99.90 	&	99.95 	&	95.80 	&	94.43	&	99.76	&	99.67	&	\textbf{100.00} 	\\
\#4 &   	90.27	&	93.54 	&	83.01	&	82.18	&	83.47	&	93.55 	&	\textbf{95.20} 	&	93.64 	&	91.53	&	91.85	&	94.88	&	95.05	\\
\#5 &    	89.36	&	82.42 	&	89.01	&	93.62	&	92.87	&	87.02 	&	84.60 	&	84.41 	&	\textbf{93.91} 	&	88.57	&	84.76	&	93.74	\\
\#6 &    	95.33	&	92.23 	&	94.15	&	98.98	&	99.35	&	95.11 	&	95.65 	&	93.87 	&	93.14	&	97.39	&	95.71	&	\textbf{99.72} 	\\
\#7 &    	96.68	&	95.23 	&	97.27	&	99.23	&	99.12	&	99.17 	&	95.72 	&	83.06 	&	96.19	&	\textbf{99.57} 	&	91.96	&	99.17	\\
\#8 &    	96.67	&	97.05 	&	93.77	&	95.94	&	97.93	&	98.70 	&	98.40 	&	99.60 	&	96.32	&	97.33	&	\textbf{100.00} 	&	98.27	\\
\#9 &   	94.61	&	93.10 	&	92.00	&	94.90	&	94.51	&	95.30 	&	95.20 	&	95.46 	&	\textbf{97.06} 	&	94.34	&	96.68	&	96.54	\\
\#10 &  	93.70	&	99.00 	&	95.04	&	94.75	&	95.59	&	94.07 	&	97.91 	&	96.67 	&	93.56	&	94.05	&	97.73	&	\textbf{98.08} 	\\
\#11 &  	92.48	&	89.82 	&	88.68	&	88.08	&	91.23	&	91.91 	&	92.61 	&	\textbf{100.00} 	&	97.73	&	99.28 	&	98.74	&	98.88	\\
\#12 &  	87.45	&	90.92 	&	96.21	&	94.33	&	95.69	&	93.60 	&	92.38 	&	99.42 	&	96.76	&	95.36	&	\textbf{99.87} 	&	99.12	\\
\#13 &   	98.88	&	99.78 	&	99.18	&	97.84	&	99.53	&	99.65 	&	\textbf{100.00} 	&	99.80 	&	99.85	&	96.31	&	99.33	&	99.91\\
\#14 &   	95.55	&	95.29	&	95.05	&	93.44	&	94.15	&	95.62 	&	93.69 	&	95.94 	&	91.28	&	95.45	&	95.19	&	\textbf{96.15}	\\
Average &   94.52   &   93.60   &	94.07   &   94.98 	&   95.03   &   95.41 	&   95.21   &   94.93   &   95.49   &   95.68   &   95.88   &   \textbf{98.11} \\

\bottomrule
	\end{tabular}
	}
\end{table*}

\todo{To evaluate the emotion recognition performance, we take four supervised learning methods (i.e., SVM, SMPM, RVM and PCVM) using all features as baselines and also compare PFCVM$_\mathit{LP}$ with seven state-of-the-art feature selection algorithms: mRMR \citep{mrmr2005}, TRC \citep{trc},
FSNM \citep{fsmn}, L$_1$SVM \cite{l1-svm}, WSVM \cite{wsvm}, JCFO \cite{krishnapuram2004}, and PFCVM$_{\mathit{EM}}$ \cite{li2014}.
Among the algorithms considered, mRMR, TRC and FSNM are filter feature selection algorithms,
L$_1$SVM, WSVM, JCFO and PFCVM$_\mathit{EM}$ are embedded feature selection algorithms.
For mRMR, TRC, and FSNM, the PCVM classifier is used to evaluate their emotion recognition performance by using the features selected by them.
In these experiments, the Gaussian RBF is used as the basis function.
Two popular evaluation criteria, i.e., error rate (ERR\footnote{$\text{ERR} = 1-\text{classification accuracy}=\frac{1}{N}\sum_{i=1}^N 1(y_i \neq f(\mathbf{x}_i; \mathbf{w},\boldsymbol \theta)) $.}) and area under the curve of the receiver operating characteristic (AUC) are adopted for evaluation; they represent a probability criterion and a threshold criterion, respectively~\citep{caruana2004data}.}

\todo{
We follow the procedure in \citep{chen2009} to choose the  parameters.
More precisely, the dataset for the 15th subject is chosen for cross-validation, in which we train each algorithm with all parameter candidates and then choose the parameters with the lowest median error rate on this dataset.
We follow this procedure to choose the optimal numbers of clusters for SMPM, the kernel parameter $\vartheta$ for RVM, SMPM, PCVM and WSVM, and the regularization parameters for JCFO and L$_1$SVM.
For SVM, the regularization $C$ and the kernel parameter $\vartheta$ are tuned by grid search, in which we train SVM with all combinations of each candidate $C$ and $\vartheta$, then choose
the combination with the lowest median error rate.
For mRMR, TRC, and FSNM, the proper sizes of feature subsets and the kernel parameter $\vartheta$ for PCVM are
chosen by a similar grid search.
We also choose the proper starting points for PFCVM$_{\mathit{EM}}$ and the initial hyperparameters for
PFCVM$_{\mathit{LP}}$.
}

\todo{For each subject, all algorithms are separately run on three sessions data,
and the results are averaged over the three sessions, which are reported in Table~\ref{eeg_classification}.
In terms of the classification error rate, PFCVM$_{\mathit{LP}}$ outperforms
all the other methods on $5$ out of $14$ datasets and also achieves competitive performance on the other datasets.
Especially, on the subject 3 dataset, PFCVM$_{\mathit{LP}}$ achieves a 3.24\%--10.74\% improvement
compared over other methods.
In terms of AUC, PFCVM$_{\mathit{LP}}$ outperforms others on $5$ datasets,
JCFO, WSVM and, FSNM win on 2 datasets, respectively.  PFCVM$_{\mathit{EM}}$, L$_1$SVM and TRC only win on 1 dataset.
We compute the average classification error rate and AUC over all datasets for each method. On average, PFCVM$_{\mathit{LP}}$ consistently outperforms all the other methods on the emotional
EEG datasets.
Compared with the other methods, PFCVM$_{\mathit{LP}}$ obtains 3.32\%--6.30\% and 2.33\%--4.82\% relative improvements for classification accuracy (1 $-$ error rate) and AUC, respectively.}

In order to give a comprehensive performance comparison between PFCVM$_{\mathit{LP}}$ and other methods with statistical significance,
the Friedman test \citep{statistical} combining with the post-hoc tests is used to make statistical comparisons of multiple methods over multiple
data sets.
The performance of two methods is significantly different if their average
ranks 
on all datasets differ by at least the critical difference:
\begin{equation}
\mathit{CD}=q_{\alpha }\sqrt{\frac{p(p+1)}{6N}},
\end{equation}
where $p$ is the number of algorithms, $N$ is the number of datasets, $\alpha$ is the significance level, and $q_{\alpha}$ denotes the critical value.
According to the experimental results in Table~\ref{eeg_classification}, the better embedded feature selection algorithms, JCFO and PFCVM$_{\mathit{EM}}$,
the better filter feature selection algorithms, TRC and FSNM, and the important baseline, PCVM, are chosen to compare with PFCVM$_{\mathit{LP}}$.
Choosing $\alpha=0.05$ and $q_{\alpha}=2.576$ ($p=6$), the critical difference becomes $\mathit{CD}=1.82$.

\begin{figure}[t]
\centering%
\subfigure[Average ranks of ERR]{
    \label{rank_err}
    \includegraphics[clip,trim=0mm 10mm 0mm 0mm,height=4cm]{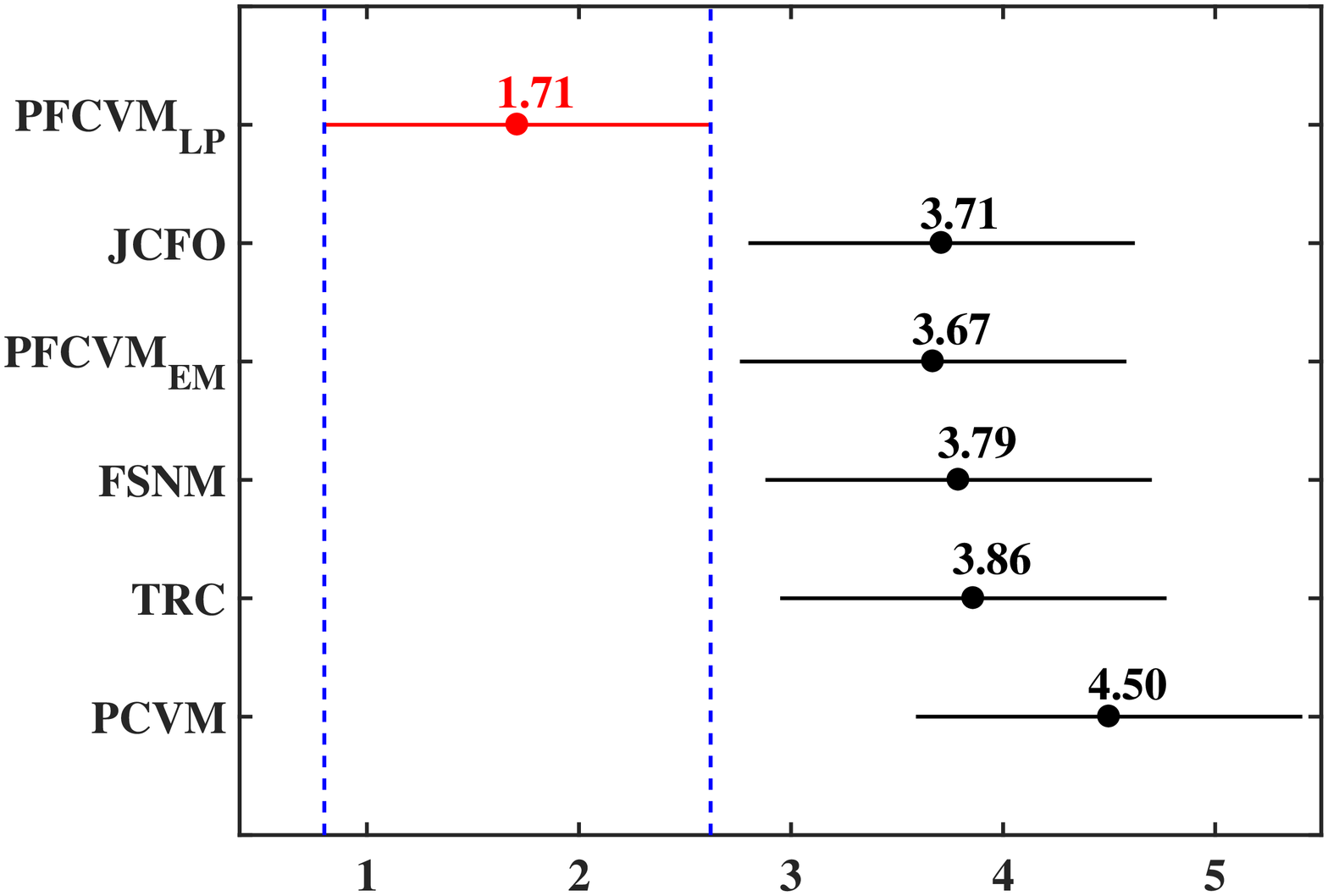}}
\subfigure[Average ranks of AUC]{
    \label{rank_auc}
    \includegraphics[clip,trim=0mm 10mm 0mm 0mm,height=4cm]{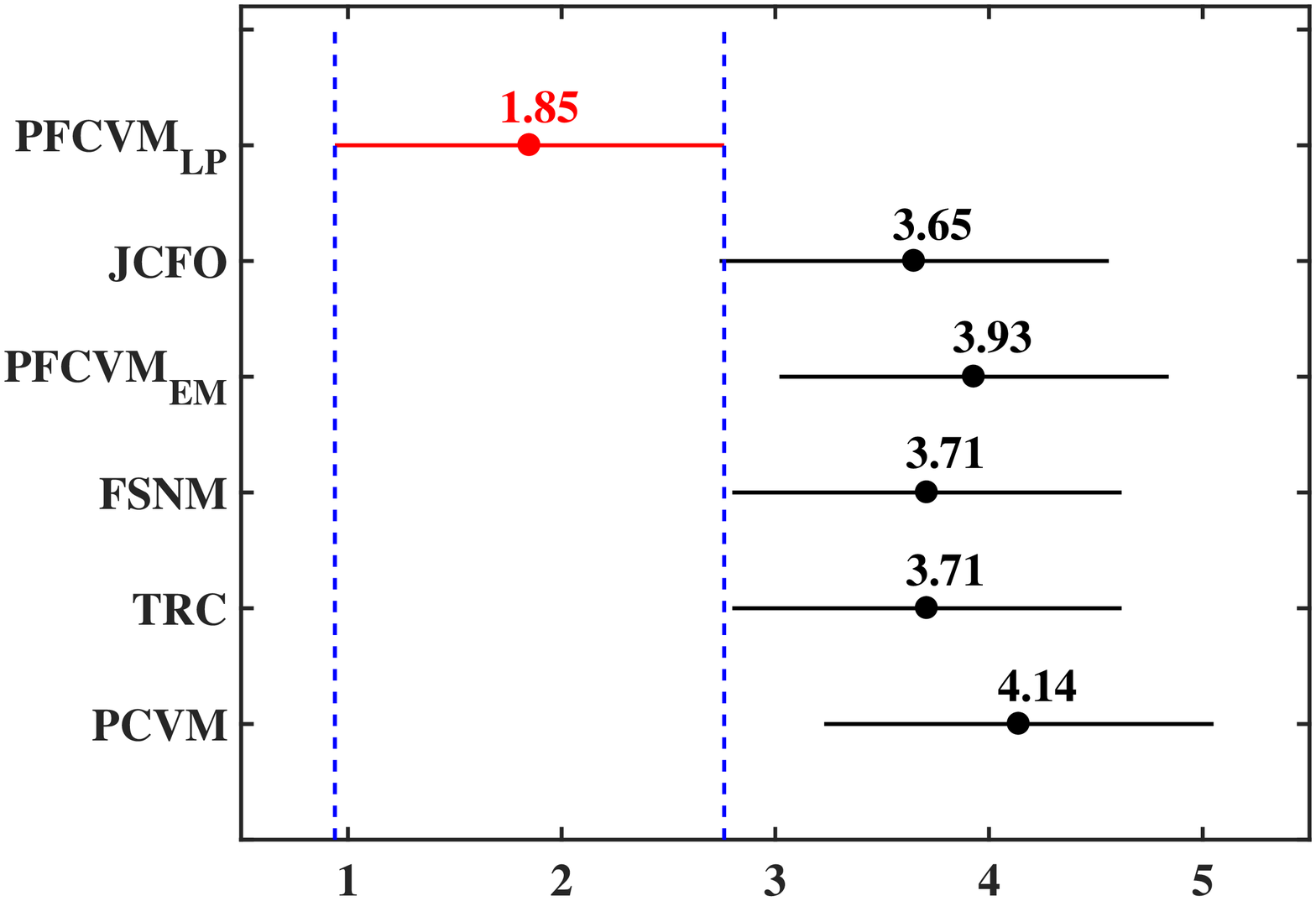}}
 \caption{Results of the Friedman test for the performance of
PFCVM$_{\mathit{LP}}$ and other algorithms on the EEG datasets.
The dots denote the average ranks, the bars indicate the critical differences CD, and the algorithms having non-overlapped bars are significantly inferior to PFCVM$_{\mathit{LP}}$.}
\label{friedman_test}
\end{figure}

\todo{Fig.~\ref{friedman_test} shows the Friedman test results on the EEG datasets. We observe that the differences between PFCVM$_{\mathit{LP}}$ and PCVM, are significant (greater than $\mathit{CD}$).
This observation is meaningful because it demonstrates the effectiveness of simultaneously learning feature weights in terms of improving performance.
We note that the differences between PFCVM$_{\mathit{LP}}$ and PFCVM$_{\mathit{EM}}$ are greater than $\mathit{CD}$, which indicates that fully Bayesian estimation approximated by the type-\uppercase\expandafter{\romannumeral2} maximum likelihood approach works better than the EM algorithm based on MAP point estimation.
Moreover, PFCVM$_{\mathit{LP}}$ achieves a significant difference compared with other feature selection algorithms (i.e., TRC, FSNM and JCFO), which demonstrates the effectiveness and superiority of PFCVM$_{\mathit{LP}}$ for selecting relevant features.}

\todo{To quantitatively assess the reliability of our classification results, the kappa statistic \cite{kappa} is adopted to evaluate the consistency between the prediction of algorithms and the truth.
The kappa statistic can be used to measure the performance of classifiers, which is more robust than classification accuracy.
In this experiment, the kappa statistic of all classifiers ranges from 0 to 1, where 0 indicates the chance agreement between the prediction and the truth, and 1 represents a perfect agreement between them.
Therefore, a larger kappa statistic value means that the corresponding classifier performs better.
Table~\ref{eeg_kappa} reports the kappa statistic for PFCVM$_{LP}$ and other methods on the emotional EEG datasets.
The standard error interval with two-sided 95\% confidence level of PFCVM$_{LP}$ is also reported in Table \ref{eeg_kappa}, which constitutes the confidence interval with the kappa statistic.
The kappa statistics of other methods lying in the confidence interval of PFCVM$_{LP}$ are marked with *, which indicates that they are not significantly worse or better than PFCVM$_{LP}$.
From Table \ref{eeg_kappa}, we observe that PFCVM$_{\mathit{LP}}$ achieves the best kappa statistics on $5$ out of $14$ datasets.
Although other methods achieve the best kappa statistics on the rest datasets, they are not significantly better than PFCVM$_{LP}$ since the best kappa statistics lie within the confidence interval of PFCVM$_{LP}$ except on the subject 8 dataset.
Overall, PFCVM$_{LP}$ is the most frequent winner in terms of kappa statistic.}

\begin{table*}[t]
	\caption{The Kappa statistic of PFCVM$_{LP}$ and other competing algorithms on the emotional EEG
datasets. The best result for each dataset is illustrated in boldface.}
	\label{eeg_kappa}\centering
	\resizebox {5.4in }{!}{
		\setlength{\tabcolsep}{0.08in}
		\begin{tabular}{ccccccccccccc}
			\toprule
Subject  &  SMPM  & PCVM & mRMR & TRC  & FSNM  & L$_1$-SVM & WSVM &   PFCVM$_{EM}$ & JCFO & PFCVM$_{LP}$    \\ \midrule
\#1 	&	83.38	&		82.51	&\ \	90.20*	&	\ \ \textbf{90.93}*	&	\ \ 90.45*	&	82.92	& \ \	89.75*	&	\ \ 89.28*	&	74.82	&	90.35	$\pm$	0.028  \\
\#2 	&	48.75	&		72.80	&	47.18	&	54.08	&	59.98	&	63.15	&	74.47	&	71.97	&	68.25	&	\textbf{78.55}  	$\pm$	0.038 \\
\#3 	&	85.06	&		81.34	&	81.26	&	87.09	&	93.67	&	85.96	&	87.08	&	92.56	&	93.28	&	\textbf{100.00}  	$\pm$	0.000 \\
\#4 	&	62.49	&		62.90	&	46.96	&	45.07	&\ \	78.79*	&	73.65	&	74.47	&	74.37	&	55.80	&	\textbf{78.88}  	$\pm$	0.031\\
\#5 	&	52.91	&		63.62	&	68.51	&	68.40	&	51.84	&	63.42	&\ \	\textbf{79.76}*	&	74.76	&	66.73	&	78.78	$\pm$	0.033 \\
\#6 	&	66.55	&		84.58	&\ \	\textbf{90.63}*	&	74.70	&	78.36	&	66.71	&	78.43	&	84.09	&	66.47	&	89.60	$\pm$	0.026 \\
\#7 	&	60.20	&\ \	\textbf{84.25}*	&	78.65	&	80.75	&	71.86	&	58.94	&\ \	80.79*	&\ \	83.97*	&	80.25	&	83.99	$\pm$	0.035 \\
\#8     &	90.10	&		82.13	&	78.41	&	92.85	&	92.77	&	94.25	&	92.02	&	86.39	&	\textbf{100.00}	&	96.90	$\pm$	0.014 \\
\#9 	&	75.18	&		75.22	&	74.23	&	70.34	&	79.82	&\ \	86.43*	&\ \	\textbf{87.60}*	&	75.97	&\ \	83.91*	&	85.01	$\pm$	0.027 \\
\#10 	&	85.98	&		76.23	&	82.56	&	76.16	&	81.53	&	81.27	&	74.97	&	82.63	&\ \	87.04*	&	\textbf{90.32}  	$\pm$	0.035 \\
\#11 	&	92.53	&		82.76	&	67.81	&	73.89	&	72.90	&\ \	\textbf{95.09}*	&	86.95	&\ \	93.86*	&\ \	93.29*	&	94.11	$\pm$	0.027 \\
\#12 	&\ \ 90.40*	&		74.47	&	77.59	&	78.57	&	76.41	&	91.78	&	86.42	&	80.16	&\ \	\textbf{92.50}*	&	91.97	$\pm$	0.018 \\
\#13 	&	90.14	&		73.72	&\ \ 96.05*	&	94.51	&\ \	95.75*	&\ \	95.75*	&\ \	96.39*	&	93.06	&	94.50	&	\textbf{96.73} 	$\pm$	0.013 \\
\#14 	& \ \ \textbf{78.78}*		&	60.88	&	63.01	&	69.32	&	60.38	&\ \	72.52*	&	70.38	&\ \	77.12*	&\ \	74.40*	&	75.02	$\pm$	0.041 \\

\bottomrule
	\end{tabular}
	}
\end{table*}

%

\begin{figure}[ht]
\centering%
\subfigure[Beta band]{
    \label{beta}
    \includegraphics[width=1.5in,height=1.5in]{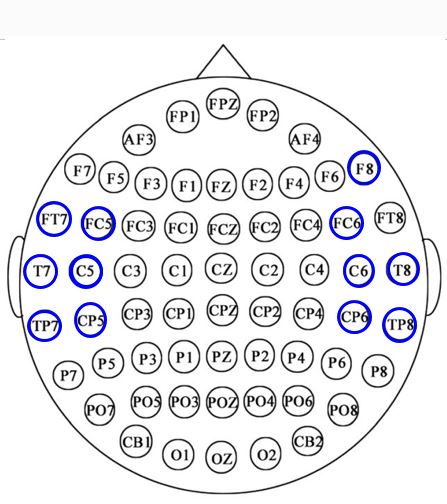}} \hspace{0.4in}
\subfigure[Gamma band]{
    \label{gamma}
    \includegraphics[width=1.5in,height=1.5in]{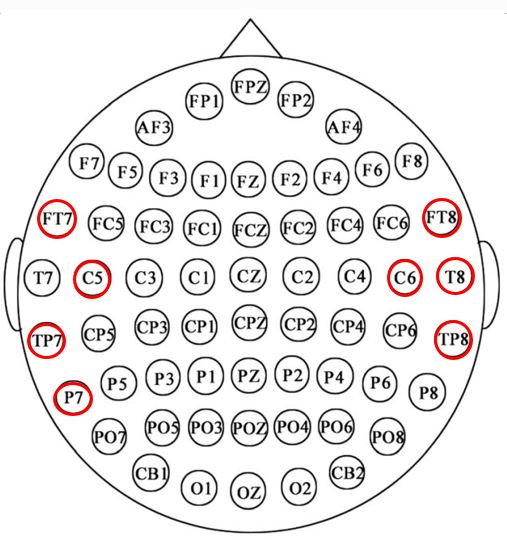}}
 \caption{Profiles of top 20 features selected by PFCVM$_{\mathit{LP}}$ on the Beta and Gamma frequency bands.}
\label{eeg_fig}
\end{figure}

According to \citet{zheng2015}, there are several irrelevant EEG channels in the SEED dataset, which will introduce noise to emotion recognition, and degrade the performance of classifiers.
To illustrate the ability of PFCVM$_{\mathit{LP}}$ for selecting discriminative features and remove irrelevant features, Fig.~\ref{eeg_fig} illustrates the positions of the top 20 features selected by PFCVM$_{\mathit{LP}}$.
As depicted in the figure, the top 20 features are all from the Beta and Gamma frequency bands and located at the lateral temporal area, which is consistent with previous findings~\citep{zheng2015, zheng2017}.
This result indicates that PFCVM$_{\mathit{LP}}$ can effectively select the relevant channels containing discriminative information and simultaneously eliminate irrelevant channels for emotion recognition task.

\subsection{High-dimensional gene expression data: Performance in the presence of many irrelevant features}
\label{sec:exp:gene}

\todo{Gene expression datasets contain lots of irrelevant features, which  may degrade the performance of classifiers~\citep{colon,golub1999Leukemia,krishnapuram2003}.
In this experiment, three gene expression datasets: colon cancer \citep{colon}, Duke cancer \citep{duke2001predicting}, and ALLAML \citep{golub1999Leukemia} are chosen to examine whether PFCVM$_{\mathit{LP}}$ is able to eliminate the irrelevant features and make informative predictions for high-dimensional data.
The colon cancer dataset includes expression levels of $2,000$ gene features from $62$ different samples, in which $40$ samples are tumor colon and $22$ normal colon tissues.
The Duke cancer dataset contains expression levels of $7,129$ genes from $42$ tumor samples, in which $21$ samples are estrogen receptor-positive tumors and the rest of the samples are estrogen receptor-negative tumors.
The ALLAML dataset comes from $47$ normal and $40$ cancer tissues.
The ALLAML dataset consists of $72$ samples in two classes, acute lymphoblastic leukemia (ALL) and acute myeloid leukemia (AML), which have $47$ and $25$ instances, respectively.
Each sample is represented by $7,129$ gene expression values.}

Considering the relatively small numbers of samples versus the large numbers of features, in this set of experiments, we use the leave one out cross validation (LOOCV) method: each time a sample is left out to be diagnosed, and the classification model is trained to fit the remaining data.
So, we generate $62$, $42$ and $72$ runs for the colon cancer dataset, the Duke cancer dataset, and the  ALLAML dataset, respectively.
Following \citep{shevade2003simple}, in preprocessing each dataset is normalized in two ways:  sample-wise to follow a standard normal distribution and then dimension-wise to follow a standard normal distribution.

We compare PFCVM$_{\mathit{LP}}$ with PCVM and three filter feature selection algorithms, including mRMR, TRC and,
FSNM, with PCVM as the classifier.
As before, we also compare PFCVM$_{\mathit{LP}}$ with other feature and classifier co-learning algorithms,
i.e., L$_1$SVM, WSVM, JCFO, and PFCVM$_{\mathit{EM}}$.
As the dimensions of gene data are relatively high, we choose the inner product, i.e., the linear kernel, as the metric.
Finally, we report the averages of all runs on each dataset as the results, i.e., the averages of $62$, $42$ and $72$ runs on the colon
cancer, Duke cancer, and  ALLAML datasets, respectively.

\begin{figure*}[ht]
	\centering
		\includegraphics[clip,trim=100mm 37mm 110mm 7mm,height=5cm]{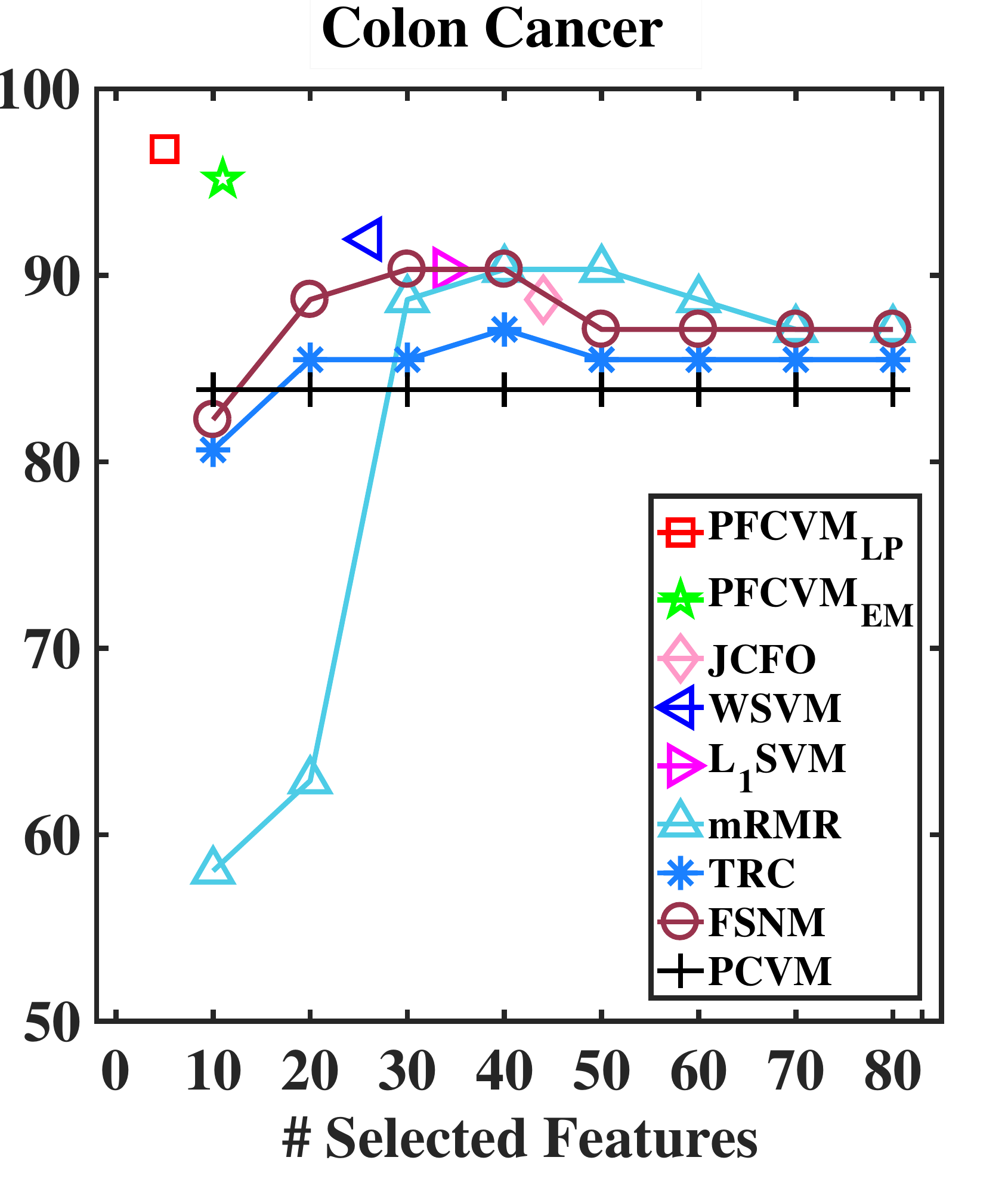}
		\includegraphics[clip,trim=100mm 0mm 40mm 20mm,height=5cm]{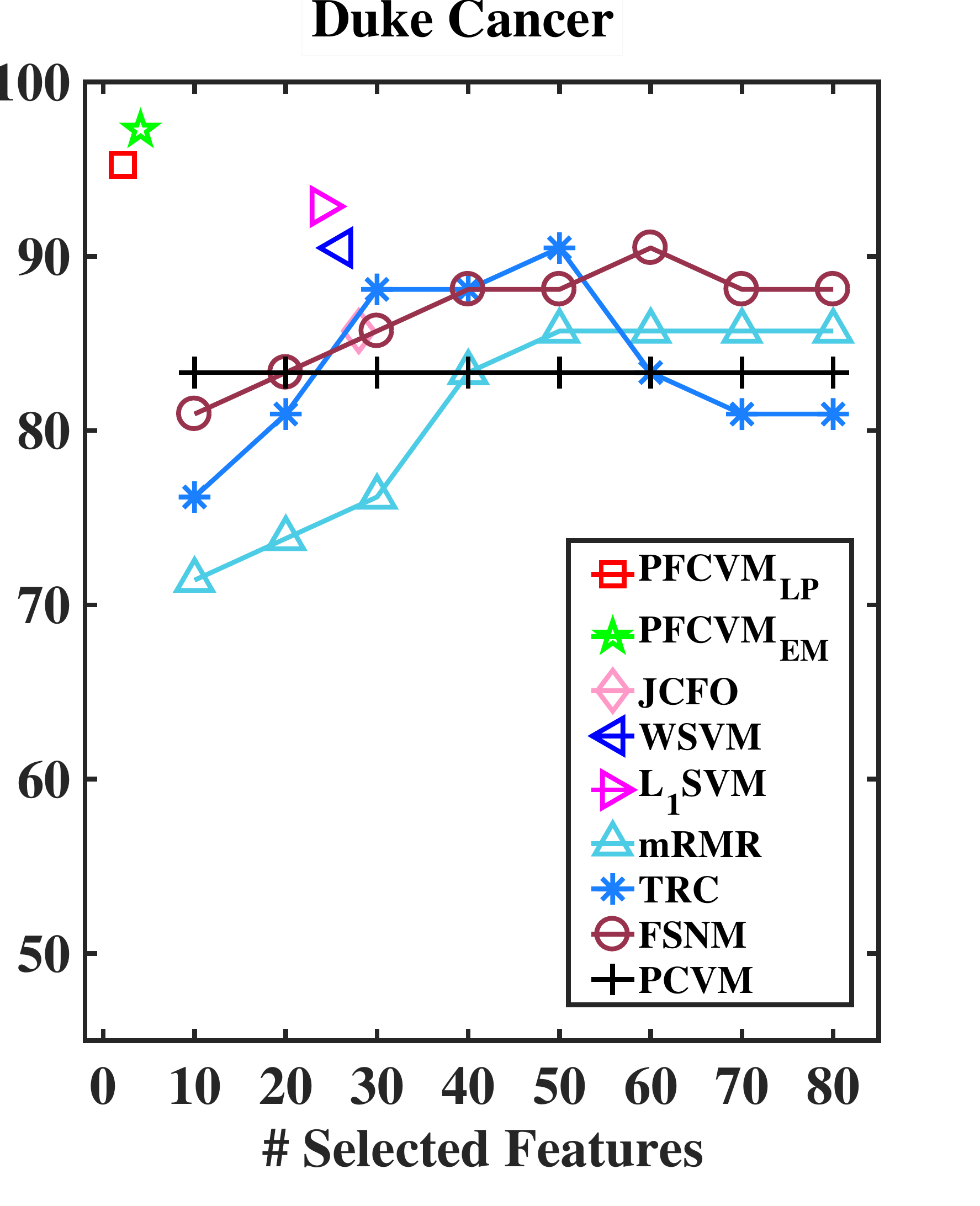}
		\includegraphics[clip,trim=5mm 10mm 20mm 0mm,height=5cm]{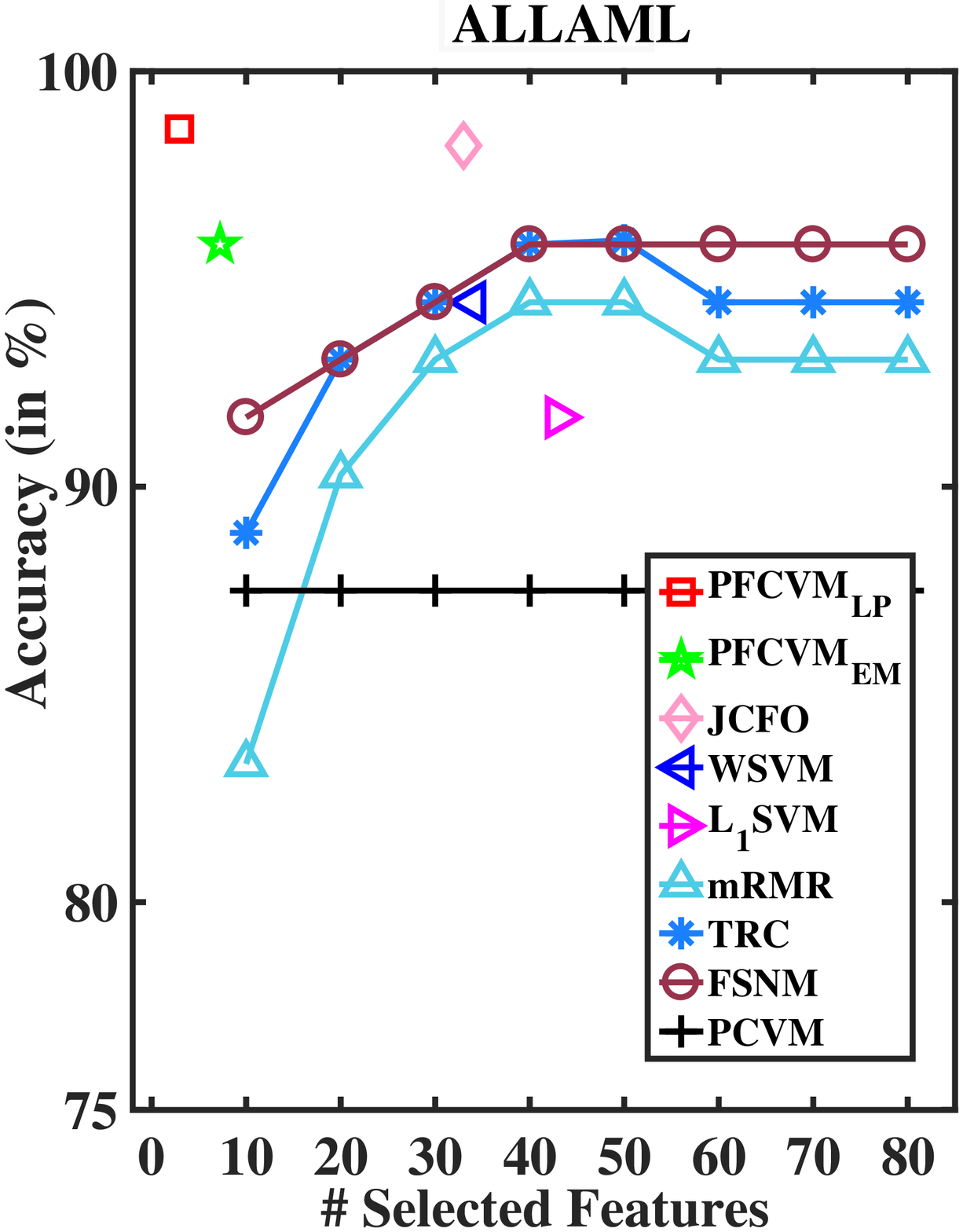}
	\caption{Accuracy curves of different algorithms with different scales of selected features. }
	\label{fig:benchmark}
\end{figure*}

\begin{figure*}[ht]
\centering
\subfigure{
   \includegraphics[width=.32\textwidth]{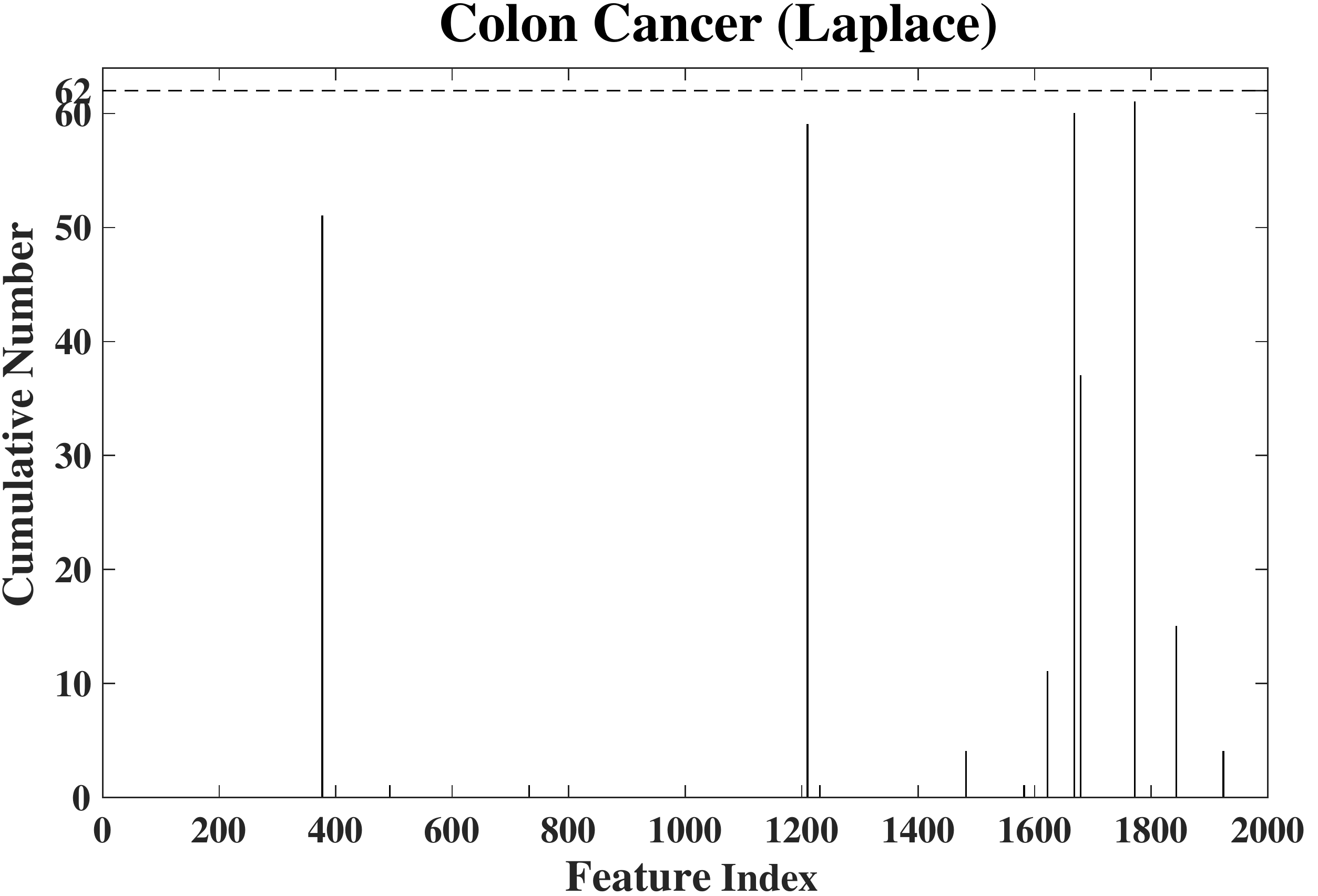}\hspace{0.02in}
   \includegraphics[width=.32\textwidth]{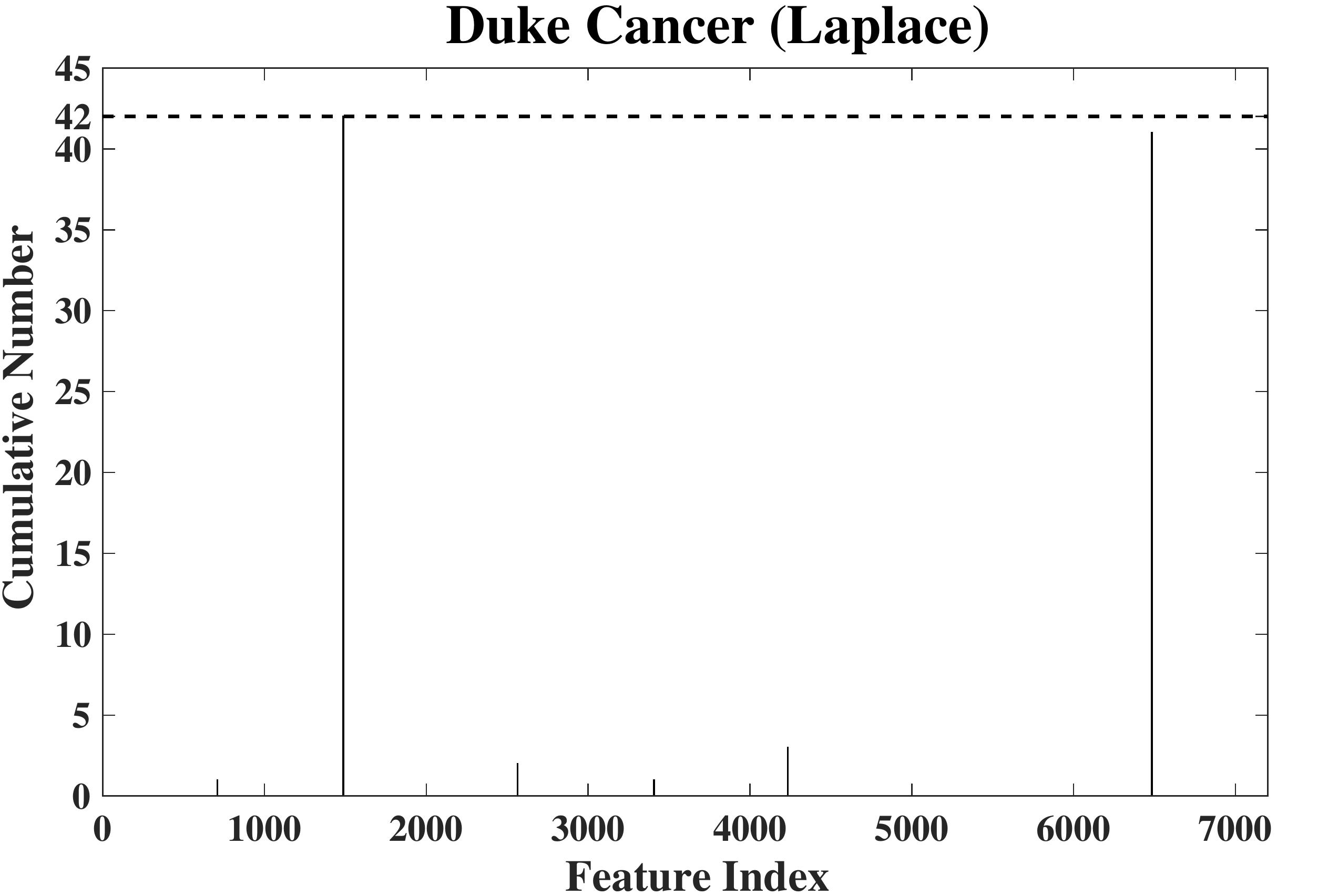}\hspace{0.02in}
   \includegraphics[width=.32\textwidth]{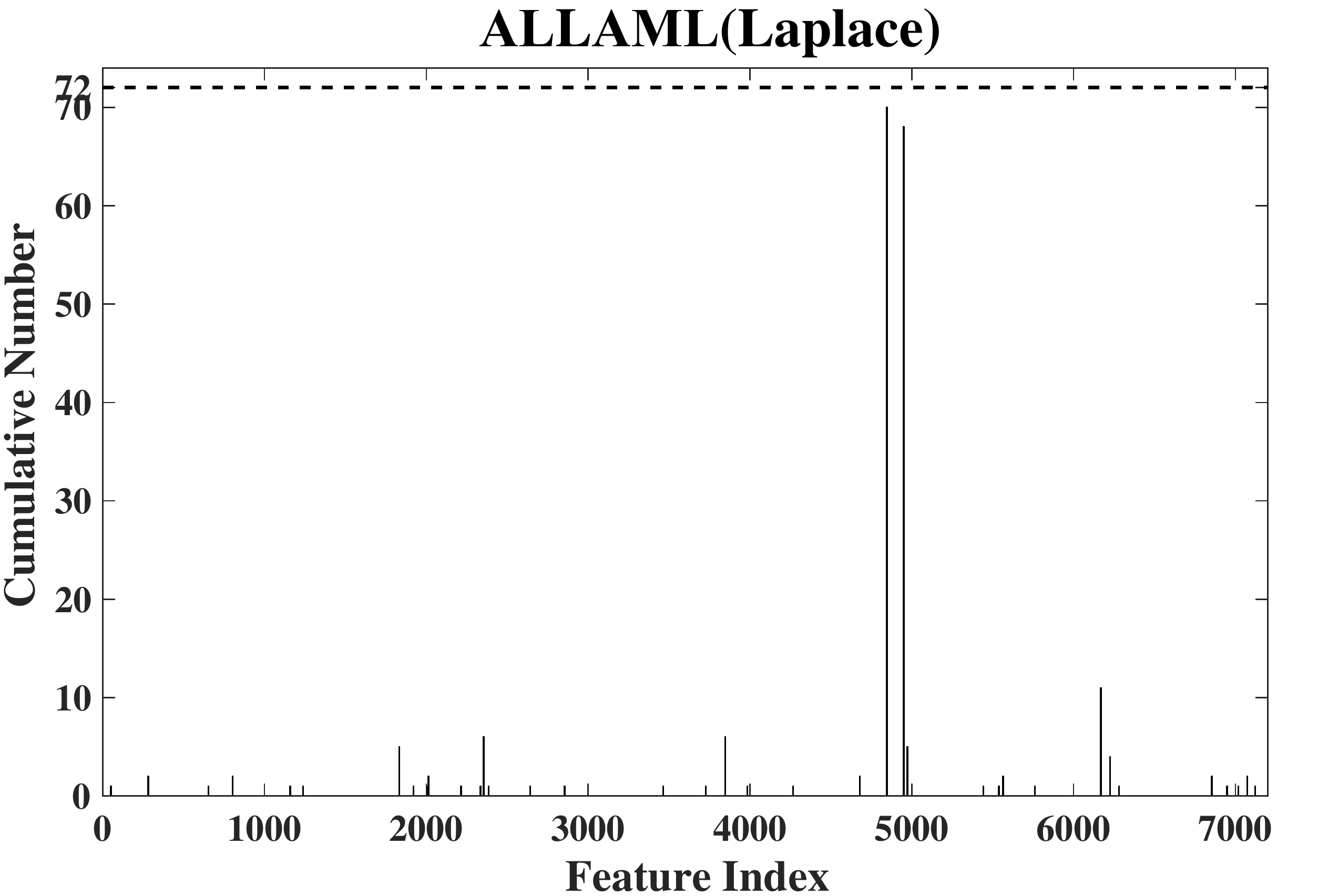}
   }
\subfigure{
    \includegraphics[width=.32\textwidth]{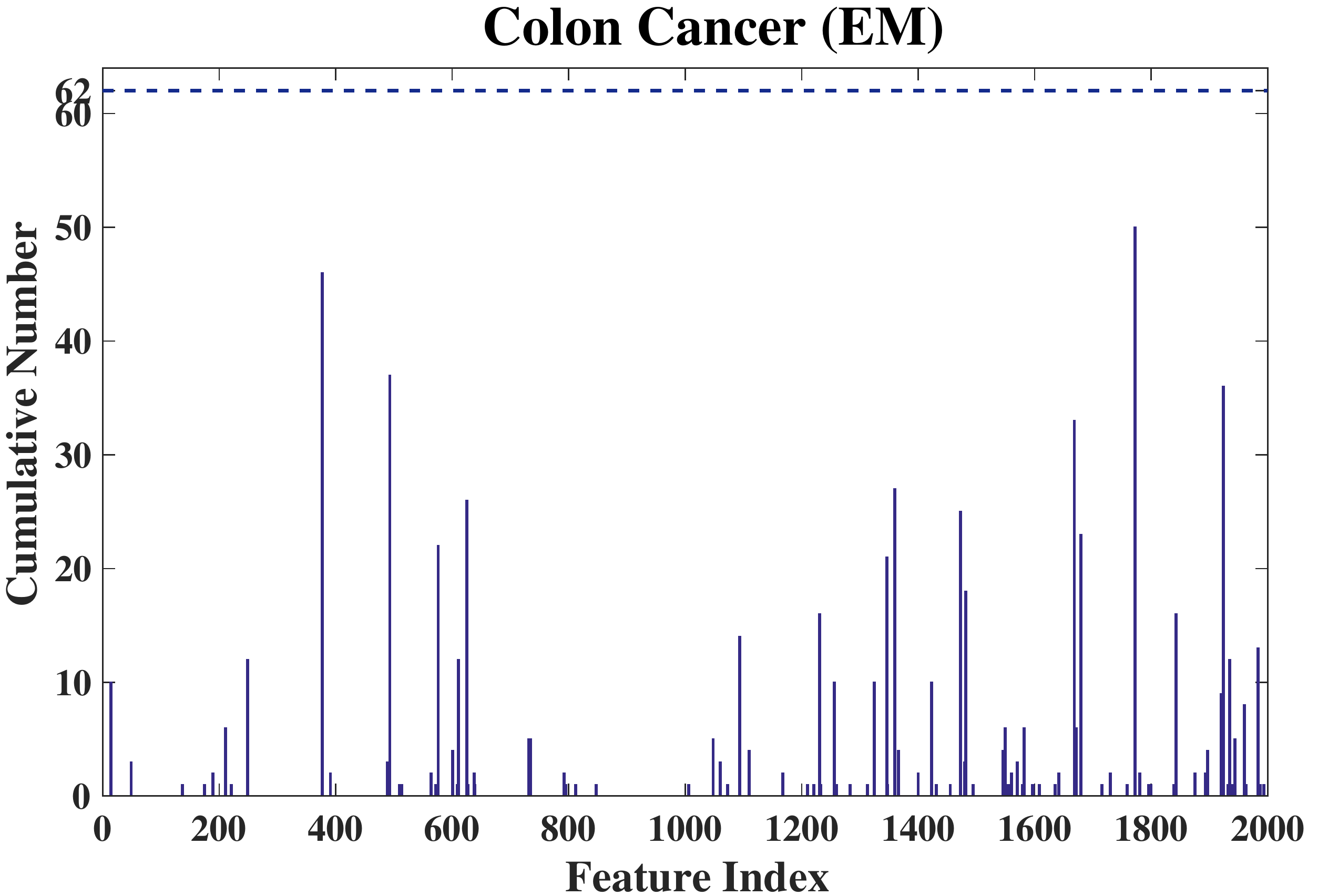}\hspace{0.02in}
    \includegraphics[width=.32\textwidth]{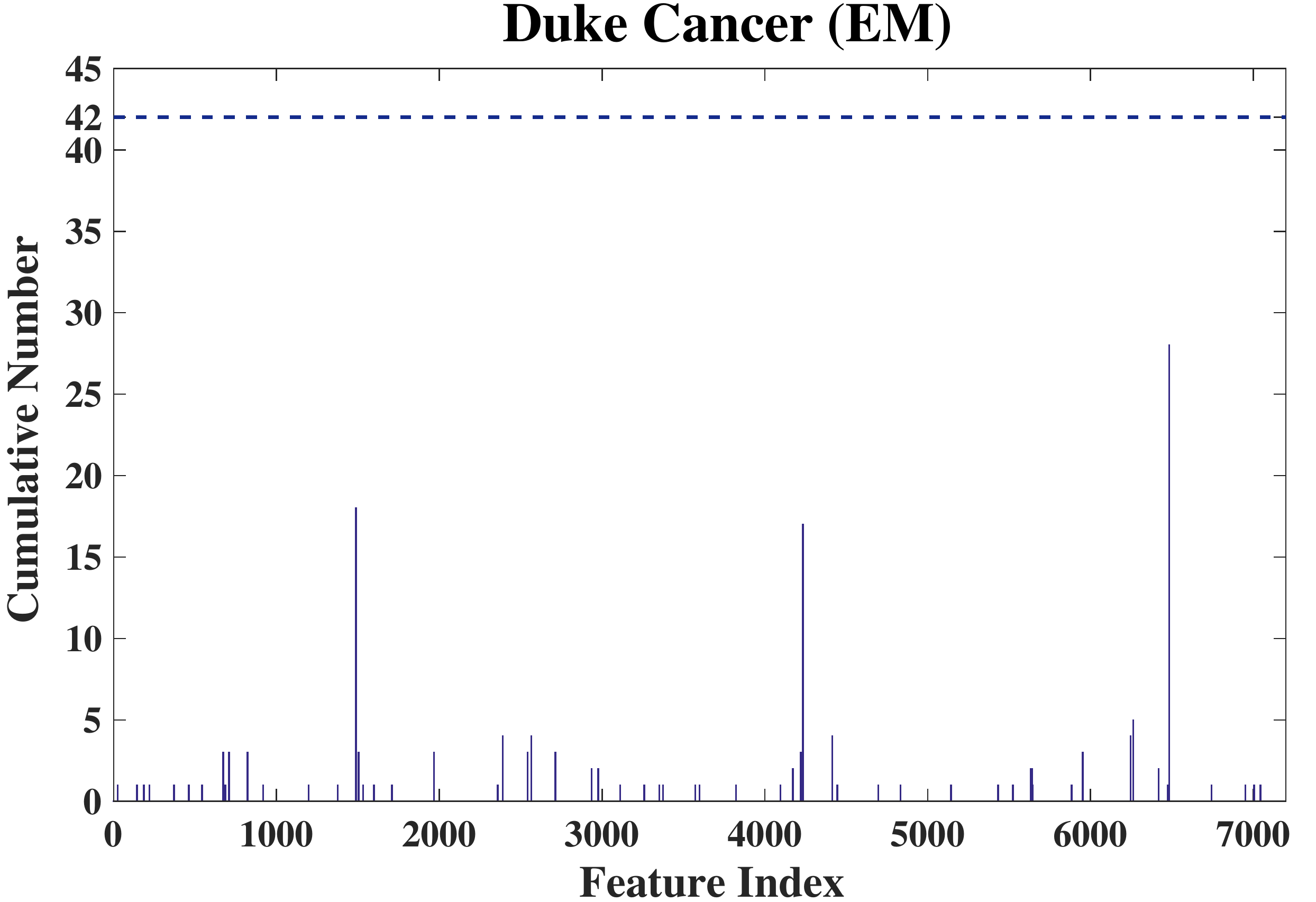}\hspace{0.02in}
    \includegraphics[width=.32\textwidth]{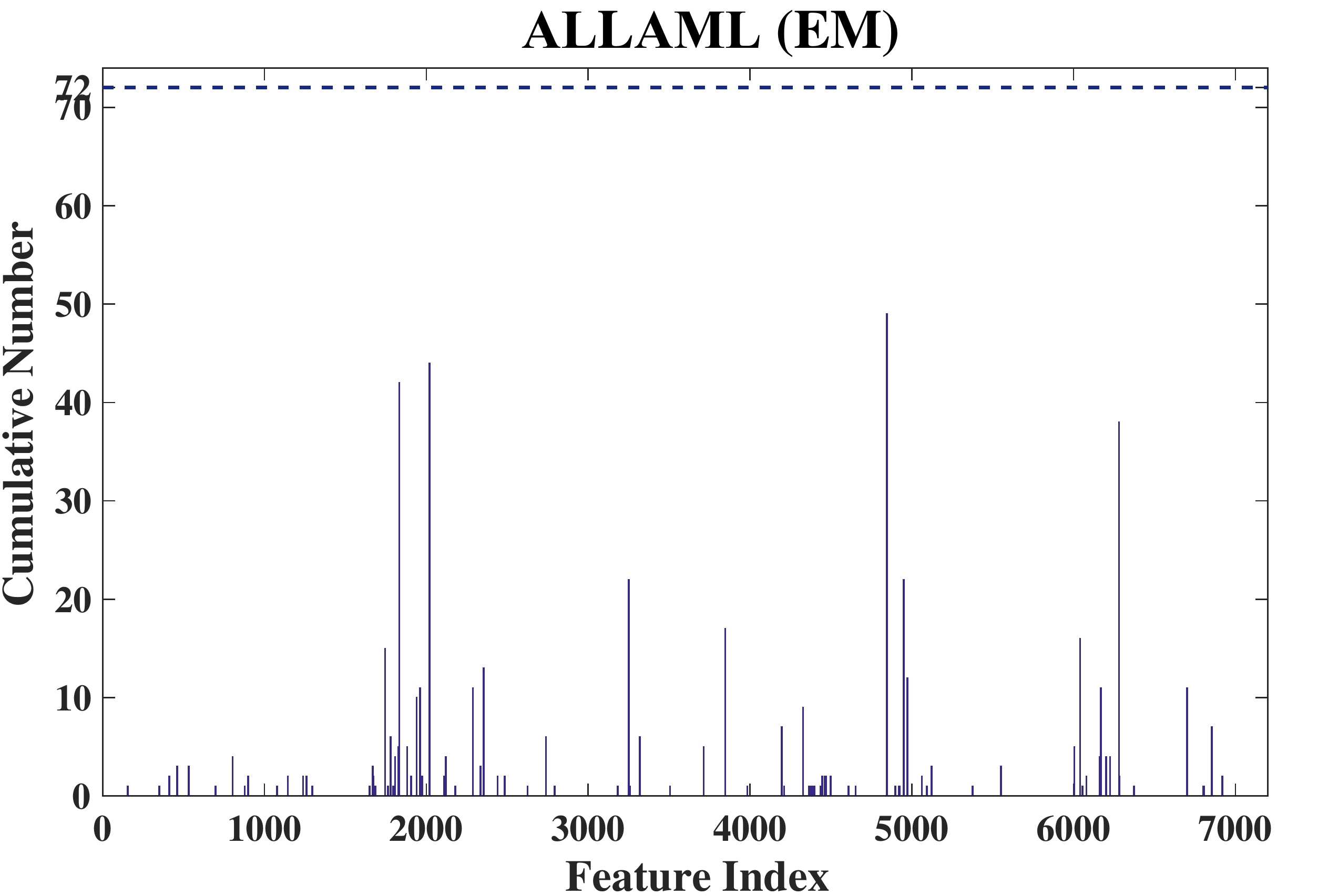}
}

\caption{Illustrations of selected features on gene expression datasets.
	 The horizontal axis shows the index of features and the vertical axis
   shows the cumulative number of occurrences for the corresponding feature. The dashed line in each figure indicates the maximum cumulative number.
   The figures at the top show the features selected by PFCVM$_{\mathit{LP}}$; those at the bottom show the features selected by PFCVM$_{\mathit{EM}}$.}
\label{fig:feature}
\end{figure*}

Fig.~\ref{fig:benchmark} shows the accuracy curves of feature selection algorithms.
For mRMR, TRC, and FSNM, the span of the selected features is $[10,20,\ldots,80]$. We train a PCVM with each number of selected features and then plot the accuracy.
Comparing our algorithms with others, from Fig.~\ref{fig:benchmark} we see that PFCVM$_{\mathit{LP}}$ achieves the highest accuracies
on $2$ out of $3$ datasets.
Moreover, on average PFCVM$_{\mathit{LP}}$ selects the smallest subsets of features on every dataset.
In Fig.~\ref{fig:feature}, we illustrate the cumulative number of occurrences\footnote{We use the phrase ``cumulative number of occurrences'' to refer to the number of times a feature is selected in the experiments, e.g., if a feature is never selected in the experiments, its ``cumulative number'' is $0$. } for each feature on each dataset with PFCVM$_{\mathit{LP}}$ and PFCVM$_{\mathit{EM}}$.
The number of features selected by the two algorithms is similar on average, but from Fig.~\ref{fig:feature}, we see that
PFCVM$_{\mathit{LP}}$ concentrates on smaller sets of relevant features than PFCVM$_{\mathit{EM}}$.
This result demonstrates that a complete Bayesian solution approximated by the type-\uppercase\expandafter{\romannumeral2} maximum
likelihood is more stable than a solution based on the EM algorithm.

\begin{table}[ht]
	\caption{Biological significance of the most frequently occurring genes in the colon cancer data set. $\#$ denotes the number of occurrences for a feature in all runs.}
	\label{tb:cancergen}
		\setlength{\tabcolsep}{0.05in}
		\centering
		\noindent
\resizebox {3.4in }{!}{
		\begin{tabular} {lllll}
			\toprule
			Feature ID& \# & GenBank ID &Description \citep{colon} \\
			\midrule
			1772 &61 &0H8393 &Collagen alpha 2(XI) chain \\
			1668 &60 &M82919 & mRNA for GABAA receptor \\
			1210 &58 &R55310 &Mitochondrial processing peptidase \\
			377  &51 &R39681 &Eukaryotic initiation factor \\
			1679 &37 &X53586 &mRNA for integrin alpha 6  \\
			\bottomrule
	\end{tabular}}
\end{table}

\begin{table}[ht]
	\caption{Biological significance of the most frequently occurring genes in the ALLAML data set. $\#$ denotes the number of occurrences for a feature in all runs. The superscript $^{*}$ denotes that these genes are  among the top $50$ most important genes for diagnosing AML/ALL \citep{golub1999Leukemia}.}
	\label{tb:leukemiagen}
		\centering
		\noindent
\resizebox {3.4in }{!}{
		\begin{tabular}{lllll}
			\toprule
			Feature ID & \# & GenBank ID & Relation & Description \citep{golub1999Leukemia} \\
			\midrule
			4847$^*$ &70 &X95735 &AML &Zyxin \\
			4951 &  68 &Y07604 &N/A &Nucleoside diphosphate  \\
			6169  &11 &M13690 &AML &Hereditary angioedema \\
			3847$^*$ &6 &U82759 &AML  &HoxA9 mRNA \\
			2354$^*$ &6 &M92287 &AML  &CCND3, Cyclin D3 \\
			4973$^*$ &5 &Y08612 &ALL &Protein RABAPTIN-5  \\
			1834$^*$ &5 &M23197 &AML & CD33 antigen \\
			\bottomrule
	\end{tabular}}
\end{table}

On the colon cancer dataset, PFCVM$_{\mathit{LP}}$ selects $4.94$ features, on average.
Among all $2,000$ genes, $5$ of them are particularly important (occurring in more than half of the tests).
The biological explanations of these $5$ genes are reported in Table~\ref{tb:cancergen} and $2$ of them (No.~$377$ and No.~$1772$) are the same genes selected by~\citet{ard2002li}.
On the Duke cancer dataset, PFCVM$_{\mathit{LP}}$ on average selects $2.14$ features and $2$ of them are selected in almost every run.
On the ALLAML dataset, PFCVM$_{\mathit{LP}}$ selects $2.94$ features on average, and $5$ of the $7$ most occurred genes are among the $50$ genes most correlated with the diagnosis \citep{golub1999Leukemia}.
The biological significance of these genes is reported in Table~\ref{tb:leukemiagen}.

\begin{table}[ht]
	\caption{Accuracy of diagnoses on gene expression datasets. }
		\setlength{\tabcolsep}{0.12in}
		\centering%
		\noindent%
\resizebox {3.4in }{!}{
		\begin{tabular}{lrrr}
			\toprule
			\bf Accuracy (\%)       &Colon Cancer   &Duke Cancer  & ALLAML \\
			\midrule
			RVM                 &85.48          &80.95        &93.06    \\
			SVM                 &83.87          &85.71        &87.50    \\
            SMPM                &75.81			&80.95		  &76.39    \\
			RVM$_{\mathit{PFCVM_{LP}}}$       &87.10          &92.86        &95.83    \\
			SVM$_{\mathit{PFCVM_{LP}}}$       &85.48          &\textbf{97.62}        &95.83  \\
            SMPM$_{\mathit{PFCVM_{LP}}}$      &86.26			&90.48		  &94.44    \\
			PFCVM$_{\mathit{LP}}$        &\textbf{96.77} &95.24        &\textbf{98.61}  \\
			\bottomrule
	\end{tabular}}
	\label{tb:gene}
\end{table}

\todo{To analyze the usefulness of the selected feature subsets by PFCVM$_{\mathit{LP}}$ for other methods, we run RVM$_{\mathit{PFCVM_{LP}}}$, SVM$_{\mathit{PFCVM_{LP}}}$ and
SMPM$_{\mathit{PFCVM_{LP}}}$ with features selected by PFCVM$_{\mathit{LP}}$ and compare them with the original RVM, SVM and SMPM.
The results are reported in Table~\ref{tb:gene}.
According to this table, using the selected features by PFCVM$_{\mathit{LP}}$, RVM$_{\mathit{PFCVM_{LP}}}$, SVM$_{\mathit{PFCVM_{LP}}}$, and SMPM$_{\mathit{PFCVM_{LP}}}$ achieve better performances  comparing to the original methods.
Even to our surprise, SVM$_{\mathit{PFCVM_{LP}}}$ outperforms PFCVM$_{\mathit{LP}}$ and obtains the best prediction on the Duke cancer dataset.
This improvement demonstrates that the feature subsets selected by PFCVM$_{\mathit{LP}}$ also work well for other methods.}


%

\subsection{Complexity analysis}
\label{sec:cc}

While computing the posterior covariance $\boldsymbol \Sigma_{\boldsymbol \theta}$ in Equation \eqref{f_covariance}, we have to derive
the negative inverse of the Hessian matrix. This derivation does not guarantee a numerically accurate result, because of the ill-condition of this Hessian matrix.
Practically, we abandon the term $\mathbf E$ in Equation \eqref{f_covariance}, so that the calculation becomes:
\begin{equation} \label{eq:covariance2}
\boldsymbol{\Sigma_{\theta}} = ( \mathbf{D^T C D} + \mathbf{B} +\mathbf{O}_{\boldsymbol{\theta}})^{-1}.
\end{equation}
In Equation \eqref{eq:covariance2}, $\mathbf{B}$ and $\mathbf{O}_{\boldsymbol \theta}$ are positive definite diagonal matrices and $\mathbf{D^T C D} $ has a quadratic form. Theoretically, the Hessian is a positive definite matrix. Nevertheless, because of machine precision, ill-condition may still occur occasionally, especially
when $\beta_k$ is very large.

In the case of large $\beta_k$, especially when $\beta_k \rightarrow \infty$,
the corresponding feature weight $\theta_k$ is restricted to a small neighborhood around $0$.
So, during the iteration, we filter out this feature from our model.
Initially, all the features are contained in the model. The main computational cost is the Cholesky decomposition in computing covariances
of posteriors, $\boldsymbol \Sigma_{\boldsymbol \theta}$ and $\boldsymbol \Sigma_{\mathbf w}$, which is $O(N^3+M^3)$. Thus the computational complexity of
PFCVM$_{\mathit{LP}}$ is the same as that of PFCVM$_{\mathit{EM}}$, RSFM~\citep{mohsenzadeh2013}, and JCFO \citep{krishnapuram2004}.

As for the storage requirements for PFCVM$_{\mathit{LP}}$, the basis function matrix $\boldsymbol{\Phi_\theta}$ needs $O(N^2)$ for storage, and in the initial training stage the covariance matrices $\boldsymbol \Sigma_{\boldsymbol \theta}$ and $\boldsymbol \Sigma_{\mathbf w}$ require $O(M^2)$ and $O(N^2)$ storage, respectively.
Therefore, the overall space complexity of PFCVM$_{\mathit{LP}}$ is $O(N^2+M^2)$, which is better than RSFM with a $O(N^2M+M^2)$ space complexity.
As iterations proceed, $N$ and $M$ are rapidly decreasing, resulting in $O(\bar{N}^3 + \bar{M}^3)$ computational complexity and $O(\bar{N}^2 + \bar{M}^2)$ space complexity, where $\bar{N} \ll N$ and $\bar{M} \ll M$.
In our experiments, $N$ and $M$ rapidly decrease to relatively small numbers in the first few iterations and the training speed quickly accelerates.

\begin{figure*}[ht]
	\centering
			\includegraphics[width=0.32\textwidth]{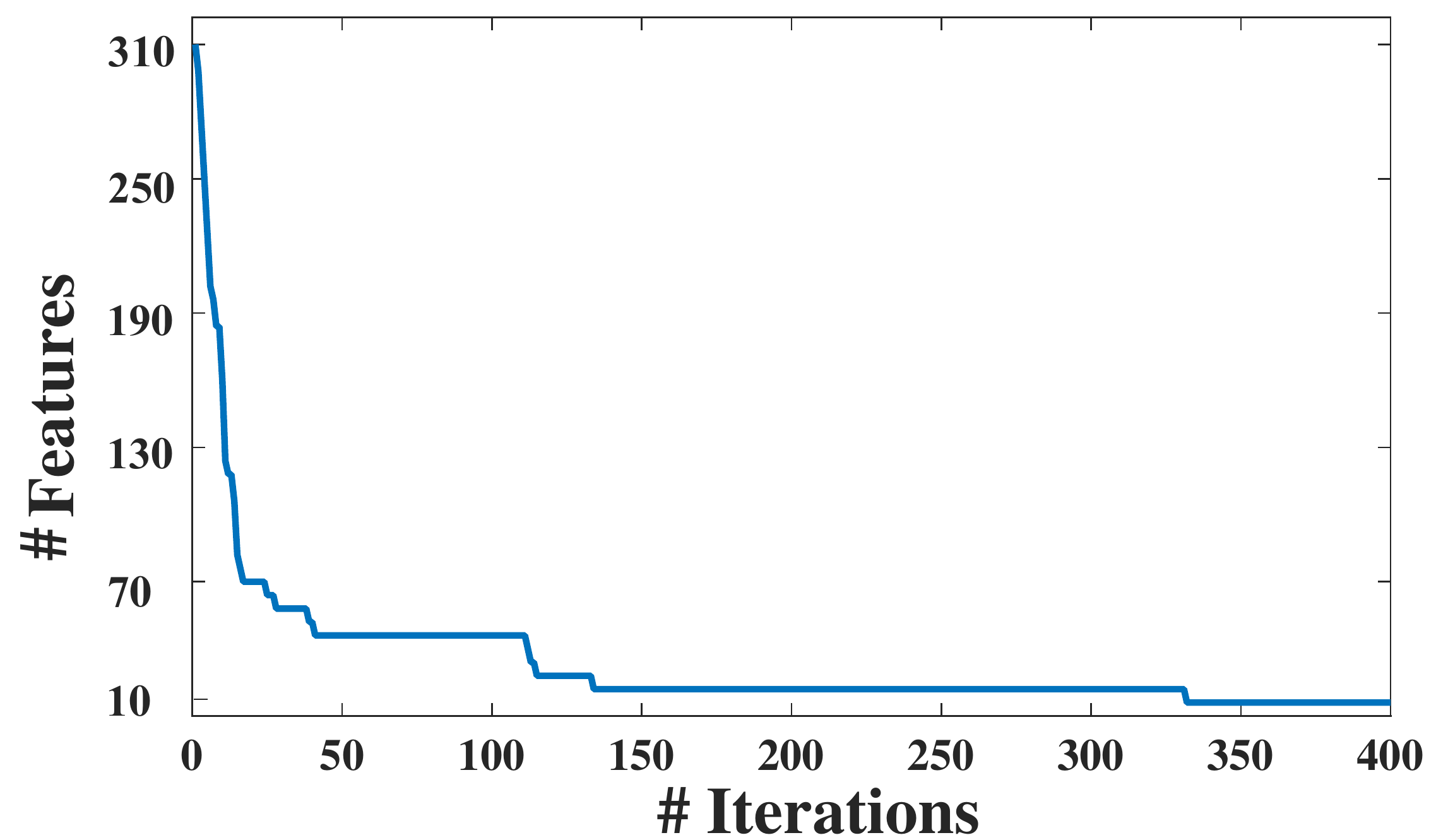}
			\includegraphics[width=0.32\textwidth]{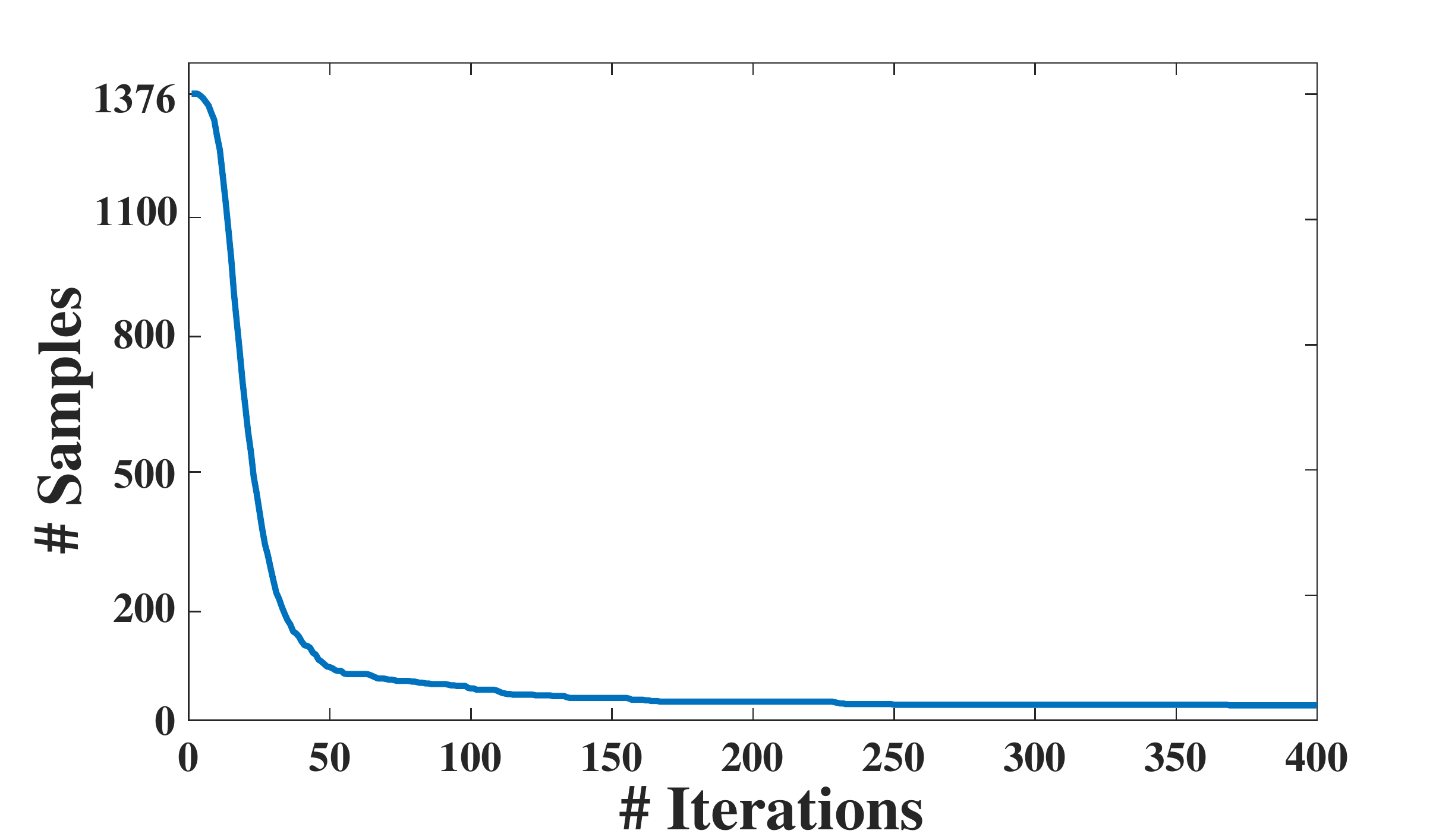}
			\includegraphics[width=0.32\textwidth]{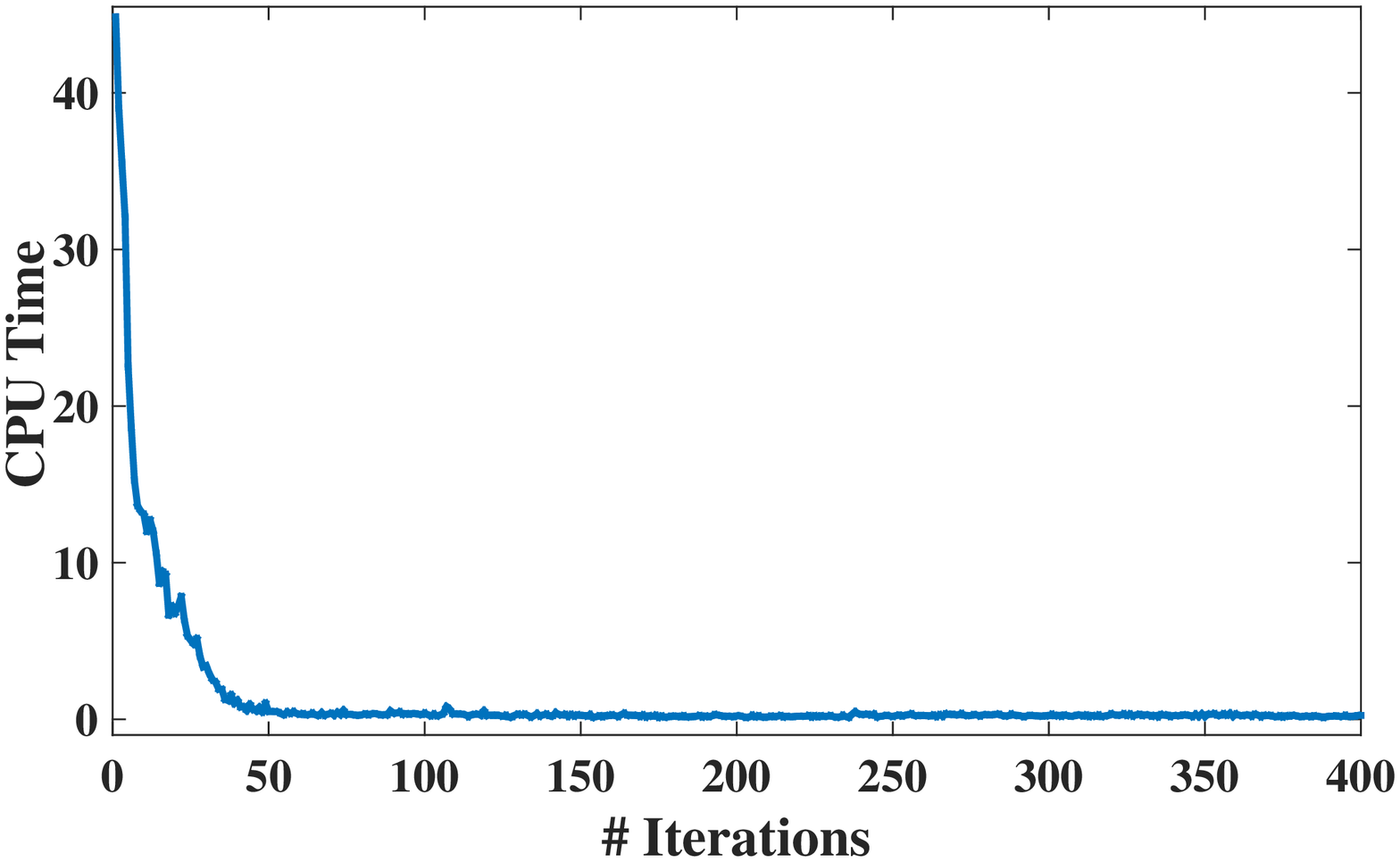}
	\caption{Illustration of the rapid decrease in the size of features, samples
		and CPU time. In these experiments, we choose subject 1 dataset as an example.
	}\label{fig:train}
\end{figure*}

As illustrated in Fig.~\ref{fig:train}, during the first $40$ iterations the size of features and samples decreases from $310$ to $40$ and from $1,376$ to $127$,
respectively, and the CPU time for each iteration step is decreased to $2.7\%$ of the first iteration.

\section{Generalization and Sparsity}
\label{sec:bound}

Both the emotional EEG and gene expression experiments indicate that the proposed classifier and feature selection co-learning algorithm is
capable of generating a sparse solution. In this section, we first analyze the KL-divergence between the prior and posterior. Following this, we investigate
the entropic constraint Rademacher complexity \citep{rbounds} and derive a generalization bound for PFCVM$_{\mathit{LP}}$. By tightening the bound, we theoretically demonstrate the significance of the sparsity assumption and introduce a method to choose the initial values for PFCVM$_{\mathit{LP}}$.

\begin{figure}[t]
	\centering
	\includegraphics[width=0.5\textwidth]{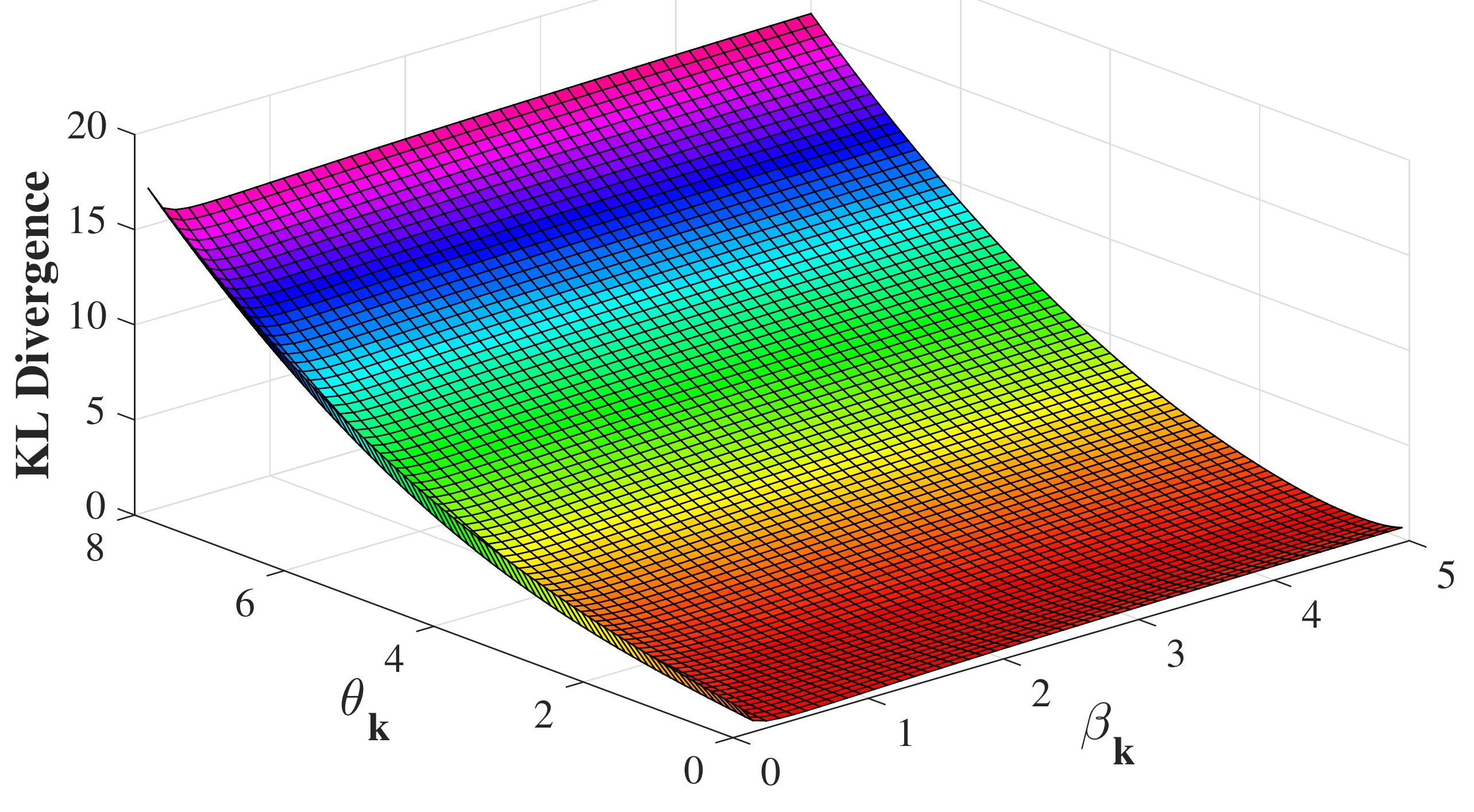}\\
	\caption{Illustration of the numerical contribution of $\boldsymbol \theta$ and $\boldsymbol \beta$  to $\mathit{KL}_{\boldsymbol \theta}(q\| p)$.
		To evaluate this contribution alone, we assume each sample weight $w_i = 1$ and each sample hyperparameter $\alpha_i = 1$.
	}\label{fig:kl}
\end{figure}

\subsection{KL-divergence between prior and posterior}
\label{sec:KLD}
In Bayesian learning, we use KL-Divergence to measure the information gain from prior to posterior.
As discussed in Section~\ref{sec:PFCVM:lap}, the approximated posterior over feature parameters, denoted as $\tilde {q}(\boldsymbol \theta) = \mathcal{N}(\boldsymbol \theta\mid \mathbf{u}_{\boldsymbol \theta}, \boldsymbol{\Sigma_{\theta}})$,
is a multivariate Gaussian distribution.
However, as the feature prior is the left-truncated Gaussian prior, the true posterior over feature parameters should
be restricted to the positive quadrant. In order to achieve this we
first compute the probability mass of the posterior in this half, $Z_0 = \int_{0}^{\infty} \tilde {q}(\boldsymbol \theta) d\boldsymbol \theta$; after that, we obtain a re-normalized version of the posterior:
$q(\boldsymbol \theta) = \tilde{q}(\boldsymbol \theta)/Z_0$, where $\theta_k \geq 0$.

We denote $\boldsymbol{\beta}_0=(\beta_{0,1}, \beta_{0,2},\ldots, \beta_{0,M}$) as the initial prior
and $\boldsymbol{\beta} = (\beta_{1}, \beta_{2}, \ldots, \beta_{M}$) as the optimized prior.
Following~\citep{chen2014}, we adopt the independent posterior assumption.
    We compute  $\mathit{KL}_{\boldsymbol \theta}(q\| p)$\footnote{$\mathit{KL}_{\boldsymbol \theta}(q\| p)$ denotes the KL-divergence between the posterior and prior in feature weights; $p$ and $q$ are short for $p(\boldsymbol{\theta}\mid \boldsymbol{\beta})$ and $q(\boldsymbol \theta) $, respectively.} using the following formula (the details are specified by \citep{choudrey2002KL}):
\begin{equation*}
\mathit{KL}_{\boldsymbol \theta}(q\| p)
=  \int_{0}^{\infty} q(\boldsymbol \theta) \frac{\ln q(\boldsymbol \theta)}{\ln p(\boldsymbol \theta\mid\boldsymbol \beta_0)}d\boldsymbol \theta 
=\sum_{k,\theta_{k}\neq 0}\left\{
\begin{array}{l}
\frac{1}{2}\left[ \frac{\beta _{0,k}}{\beta _{k}}-1+\ln \left( \frac{%
\beta _{k}}{\beta _{0,k}}\right)
 +\beta _{0,k}{\theta}_{k}^{2}\right] \\
+{\frac{(2\pi \beta _{k})^{-1/2}\left( \beta _{0,k}+\beta _{k}\right)
\theta_{k}}{\mathrm{erfcx}\left( -\,\theta_{i}\sqrt{\beta _{k}/2}\right) }} \\
-\ln \left( {\mathrm{erfc}}\left( -\frac{\theta_{k}\beta _{k}}{2}\right) \right)%
\end{array}%
\right\} ,
\end{equation*}
where $\mathrm{erfc}(z)= \frac{2}{\sqrt{\pi}}\int_z^\infty e^{-t^2}dt$ and $\mathrm{erfcx}(z) = e^{z^2}\mathrm{erfc}(z)$.

Note that $Z_{0,k} = \int_{0}^{\infty} \tilde q(\theta_k)d\theta_k = 1/2\, \mathrm{erfc}(-\theta_k\sqrt{\beta_k/2})$,
so we can calculate $\mathit{KL}_{\boldsymbol \theta}(q\| p)$ as:
\begin{eqnarray*}
\mathit{KL}_{\boldsymbol \theta}(q\| p)
 =  \sum_{k,\theta_{k}\neq 0}\left\{
\begin{array}{l}
\frac{1}{2}\left[ \frac{\beta _{0,k}}{\beta _{k}}-1+\ln \left( \frac{%
\beta _{k}}{\beta _{0,k}}\right) +\beta _{0,k}{\theta}_{k}^{2}\right] \\
+{\frac{(2\pi \beta _{k})^{-1/2}\left( \beta _{0,k}+\beta _{k}\right)
\theta_{k}}{{2\exp }\left( \beta _{k}\theta_{k}^{2}/2\right) }}Z_{0,k}^{-1}\\
-\ln \left( Z_{0,k}\right)
+\ln \left( \frac{{\mathrm{erfc}}\left( -\theta_{k}\sqrt{\beta _{k}/2}\right) }{2\,  {%
\mathrm{erfc}}\left( -\theta_{k}\beta _{k}/2\right) }\right)%
\end{array}%
\right\}.
\end{eqnarray*}
The KL-divergence is dominated by two parameters: $\boldsymbol \theta$ and $\boldsymbol \beta$. However,
the sensitivity of $\mathit{KL}_{\boldsymbol \theta}(q\| p)$ to these two parameters is different.
As shown in Fig.~\ref{fig:kl}, setting the initial hyperparameter $\beta_{0,k}=0.5$,
we see that when changing the value of $\boldsymbol \theta$, the curve of $\mathit{KL}_{\boldsymbol \theta}(q\| p)$ shows significant changes, while
this curve changes little when changing the optimized $\boldsymbol \beta_k$. Also, the minimum of $\mathit{KL}_{\boldsymbol \theta}(q\| p)$ is near the $\theta_k=0$, where
the corresponding feature is pruned.

\subsection{Rademacher complexity bound}
\label{sec:RCB}

For a binary classification learning problem the goal is to learn a function $f: \mathbb{R}^M \rightarrow \{-1,+1\}$ from a hypothesis class $F$, with the given dataset
$\mathbf S = \{x_i,y_i\}_{i=1}^N$ drawn i.i.d.\ from a distribution $D$.
We attempt to assess $f$ by the expectation loss: $L(f) = E_{(x,y)\sim D}l(y,f(x))$, where $l(y,f(x))$ is a loss function. Practically, $D$ is unaccessible and we can only assess the empirical loss for the given dataset $\mathbf S$: $\Lambda(f,\mathbf S) = \frac{1}{N}\sum_{i=1}^N l(y_i,f(x_i))$.
We adopt a 0-1 loss function: $l_{0\text{-}1}(y,f(x)) = I(yf(x)\geq0)$, where $I(\cdot)$ is the indicator function. The loss function is dominated by the $1/c$-Lipschitz function: $l_c(a) = \min(1,\max(0,1-a/c))$, namely $l_{0\text{-}1}(y,f(x)) \leq l_c(yf(x))$. Then, we conclude the entropic constraint Rademacher complexity bound in the following theorem:
\smallskip
\begin{theorem}[\citep{chen2014,rbounds}]
Based on the posterior $q(\mathbf w, \boldsymbol \theta)$ given in Section \ref{sec:PFCVM:lap}, we have the Bayesian voting classifier:
\begin{equation}
\label{eq:bvc}
\hat{y} = f(x,q) = E_{q(\mathbf w, \boldsymbol \theta)}[\mathrm{sign}(\boldsymbol{\Phi}_{\boldsymbol \theta}(\mathbf{x})\mathbf{w} )] \text{.}
\end{equation}
\end{theorem}

Define $r > 0$ and $g>0$ as arbitrary parameters. For all $f\in F$, defined at the start of Section \ref{sec:RCB}, with probability at least $1-\delta$, the bound for the generalization error of PFCVM$_{\mathit{LP}}$ on a given dataset $\mathbf S$ holds:
\begin{equation}
P(yf(x) <0)\leq  \Lambda(f, \mathbf S) + \frac{2}{c} \sqrt{\frac{2\tilde{g}(q(\mathbf w, \boldsymbol \theta))}{N}}+ \sqrt{\frac{\ln \log _{r}\frac{r\tilde{ g}(q(\mathbf w, \boldsymbol \theta))}{g }+\frac{1}{2}
\ln \frac{1}{\delta}}{N}},
\label{eq:rbound}
\end{equation}
where $c$ is the $1/c$-Lipschitz parameter, the empirical loss $\Lambda(f, \mathbf S) = \frac{1}{N}\sum_{i=1}^N l_c(y_if(x_i))$, and
the Rademacher entropic constraints $\tilde{ g}(q(\mathbf w, \boldsymbol \theta)) = r\cdot \max\{\mathit{KL}(Q\| P),g \}$. $\mathit{KL}(Q\| P)$ is the KL-divergence between the posteriors and priors in the sample and feature weights.

According to Equation (\ref{eq:rbound}), we observe that with a constant training set, the generalization error of PFCVM$_{\mathit{LP}}$ is mainly bounded by the empirical loss and $\tilde{ g}(q(\mathbf w, \boldsymbol \theta))$, in which the latter is determined by  $\mathit{KL}(Q\| P)$. Therefore, when the empirical loss is acceptable, a smaller $\mathit{KL}(Q\| P)$ could lead to a tighter bound. 
The contribution of $\mathbf w$ (the sample weight) has been analyzed in \citep{chen2014}. In order to analyze the effect of
$\boldsymbol \theta$ (the feature weight) alone, we can assume the $\mathbf w$ is a given constant value $\boldsymbol{\hat{w}}$.
As a result, we have $q(\boldsymbol{\hat{w}},\boldsymbol \theta)=q(\boldsymbol \theta\mid \mathbf {\hat w})$ and thus $\mathit{KL}(Q\| P)=\mathit{KL}_{\boldsymbol \theta\mid \boldsymbol{\hat w}} (q\| p)$.
As shown in Fig.~\ref{fig:kl}, the minimal $\mathit{KL}_{\boldsymbol \theta\mid \boldsymbol{\hat w}}(q\| p)$ is near $\theta_k$ = $0$. This is consistent with our prior assumption and demonstrates that a truncated Gaussian (sparse) prior over features can benefit the generalization performance by running as a regularization term and simultaneously encourage sparsity in feature space. Furthermore, in our model, the posterior and marginal likelihood are maximized iteratively in the training step. To accelerate the speed of convergence, we may choose proper starting points by minimizing $\mathit{KL}_{\boldsymbol \theta\mid \mathbf w}(q\| p)$, i.e., as indicated at the end of Section \ref{sec:KLD}, we can use an optimal $\boldsymbol \beta$ instead of $\boldsymbol \beta_0$ as  initial hyperparameter.

\section{Conclusion}

\label{sec:conclusion}

We have proposed a joint classification and feature learning algorithm PFCVM$_{\mathit{LP}}$.
The proposed algorithm adopts sparseness-promoting priors for both sample and feature weights to jointly learn to select the informative samples and features.
By using the Laplace approximation, we compute a complete Bayesian estimation of PFCVM$_{\mathit{LP}}$, which is more stable than previously considered EM-based solutions.
The performance of PFCVM$_{\mathit{LP}}$ has been examined according to two criteria: the accuracy of its classification results and its ability to select features.
Our experiments demonstrate that the recognition performance of PFCVM$_{\mathit{LP}}$ on  EEG emotion recognition datasets is either the best or close to the best.
On high-dimensional gene expression datasets, PFCVM$_{\mathit{LP}}$ performs more accurately when compared to other approaches.
A Rademacher complexity bound is derived for the proposed method.
By tightening this bound, we demonstrate the significance of feature selection and introduce a way of finding proper initial values.

PFCVM$_{\mathit{LP}}$ jointly encourages sparsity to features and samples.
However, in order to select features for non-linear basis functions, we have to differentiate, which leads to high computational costs.
As future work, we plan to use incremental learning \citep{chen2014,tipping2002} to reduce the computational costs.
We also plan to design an online strategy \citep{online2015} for joint feature and classifier learning.
Also, PFCVM$_{\mathit{LP}}$ focuses on the supervised binary classification. It would be interesting to extend PFCVM$_{\mathit{LP}}$ to solve multi-class problems~\citep{krishnapuram2005ml,mrvm} and semi-supervised form \citep{jiang2017}.
Finally, we aim to use PFCVM$_{\mathit{LP}}$ in other areas of research, such as in bioinformatics problems and clinical diagnoses \citep{wilkinson2007bayesian}.

\begin{acks}
We are grateful to the associate editor and the anonymous reviewers for the constructive feedback and suggestions.
\end{acks}


\section*{APPENDIXES }
\subsection*{A.\ Hyperparameter Optimization}
\label{ap:ML}

In order to compute a complete Bayesian classifier, feature and classifier co-learning includes computing this formula:
\begin{equation}
(\boldsymbol{\alpha},\boldsymbol \beta) = \arg \max_{(\boldsymbol \alpha,\boldsymbol \beta)} ~p(\mathbf{t}\mid \boldsymbol \alpha,\boldsymbol \beta,\mathbf S)p(\boldsymbol \alpha)p(\boldsymbol \beta )\text{,}
\label{eq:app:marginal}
\end{equation}
where we assume $\boldsymbol \alpha$ and $\boldsymbol \beta$ are mutually independent.
Equation~\eqref{eq:app:marginal} could be iteratively maximized between $\boldsymbol \alpha$ and $\boldsymbol \beta$. The re-estimation rules of $\boldsymbol{\alpha}$ have
been derived by \citep{chen2014}. In this appendix, we focus on deriving the  re-estimating rules for $\boldsymbol{\beta}$,
which means that we need to  compute the following equation:
\begin{equation}
\boldsymbol \beta = \arg \max_{\boldsymbol \beta} ~p(\mathbf{t\mid \boldsymbol \alpha^{\text{old}},\boldsymbol \beta},\mathbf S )p(\boldsymbol \beta )\text{.}
\label{eq:app}
\end{equation}
The hyperprior $p(\boldsymbol \beta)$ follows the Gamma distribution, $p(\boldsymbol \beta ) = \Pi_{k=1}^M Gam(\beta_k\mid c,d)$, where $c$ and $d$ are the parameters of the Gamma distribution.

As discussed in Section~\ref{sec:PFCVM:MML}, we calculating a closed form of the marginal likelihood is non-trivial. 
Using Bayesian rules, the marginal likelihood is expanded as follows:
\begin{equation}
p(\mathbf{t}\mid \boldsymbol \alpha^{old},\boldsymbol \beta,\mathbf S) = \frac{p(\mathbf{t\mid w},\boldsymbol \theta,\mathbf S)p(\mathbf{w}\mid \boldsymbol \alpha^{\text{old}})p(\boldsymbol \theta\mid \boldsymbol \beta)}{%
p(\mathbf{w,\boldsymbol \theta\mid t},{\boldsymbol \alpha^{\text{old}}},\boldsymbol{\beta})} \text{.}
\label{eq:marginal2}
\end{equation}
Applying approximate Gaussian distributions for the sample and feature posteriors, in Section~\ref{sec:PFCVM:lap}, we can obtain  $p(\mathbf{w,\boldsymbol \theta\mid t},{\boldsymbol \alpha},\boldsymbol{\beta})\approx \mathcal N(\mathbf{u}_{\boldsymbol \theta},\Sigma_{\boldsymbol \theta})* \mathcal N(\mathbf{u_\mathbf w,\Sigma_\mathbf w})$.
As a result, we maximize the logarithm of Equation \eqref{eq:app}:
\begin{eqnarray}
L &=&\log\left [ p(\mathbf{t}\mid \boldsymbol \alpha^{\text{old}} ,\boldsymbol \beta ) p(\boldsymbol \beta)\right] \notag \\
&=&\log p({\boldsymbol \theta \mid \boldsymbol \beta} )-\log {N(\mathbf{u}_{\boldsymbol \theta},\Sigma_{\boldsymbol \theta})}  + \log p(\boldsymbol \beta) +const  \notag \\
&=&\frac{1}{2}(\boldsymbol \epsilon^T \mathbf{B}^{-1} {\boldsymbol \epsilon}  +\log |\mathbf{B} | \notag-\log |{\mathbf H} + \mathbf{B}|)+\sum_{k=1}^M\left(c\log\beta_k -d\beta_k\right)+const \text{,}
\label{eq:marginal3}
\end{eqnarray}
where  $const$ is independent of $\boldsymbol{\beta}$, $\boldsymbol{\epsilon}=(\mathbf{D}^T(\mathbf{t}-\boldsymbol{\sigma})+ {\mathbf k}_{\boldsymbol \theta})$
is an $M$-dimensional vector and $\mathbf{H}=\mathbf{D}^{T}\mathbf{CD}+\mathbf{O}_{\boldsymbol \theta} - \mathbf{E}$ is an $M\times M$ matrix. Practically,
the latter two terms will disappear if we set $c = d= 0$.

To compute the optimal $\boldsymbol{\beta}$,
we first differentiate Equation (\ref{eq:marginal3}):
\begin{equation}
\label{eq:difhyper}
    \frac{\partial L}{\partial \beta_k} = -\frac{1}{2}\left(\frac{\epsilon_k^2}{\beta_k^2} -\frac{1}{\beta_k} + \frac{1}{\beta_k +  h_{k}}-2\frac{c}{\beta_k} +2d \right) \text{,}
\end{equation}
where $h_{k}$ denotes the $k$th diagonal elements of $\mathbf{H}$.
Note that $u^2_{\theta,k} = \frac{\epsilon_k^2}{\beta_k^2}$ and $\Sigma_{\theta,kk} =  \frac{1}{\beta_k + h_{k}}$, shown in Equations
\eqref{f_means} and \eqref{f_covariance}. So, setting Equation~\eqref{eq:difhyper} equal to $0$, we obtain the update formula for $\boldsymbol \beta$:
\begin{equation}
\beta_k^{\text{new}}  = \frac{2c+1}{u^2_{\theta,k} + \Sigma_{\theta, kk}+ 2d} \text{,}
\label{eq:hyperup}
\end{equation}
which is the same formula as the EM-based solution established by \citep{li2014}, and guarantees a local optimum.
However, if using the methodology of Bayesian Occam's razor reported by MacKay in \citep{mackay1992bayesian}, we derive more efficient update rules as follows:
\begin{equation}
  \beta_k^{\text{new}} = \frac{\gamma_k + 2c}{u^2_{\boldsymbol{\theta},k}+2d} \text{,}
\label{eq:hyperup1}
\end{equation}
where $\gamma_k \equiv 1- \beta_k\Sigma_{\theta,kk}$.

\bibliographystyle{ACM-Reference-Format}
\bibliography{tkdd2017}
\end{document}